\documentclass[conference]{IEEEtran}
\usepackage{times}
 
\usepackage[numbers]{natbib}
\usepackage{multicol}
\usepackage[bookmarks=true]{hyperref}
\usepackage{amsmath} 
\usepackage{graphicx} 
\usepackage{amssymb}
\usepackage{booktabs}
\usepackage{array}

\hypersetup{
    colorlinks=true,
    linkcolor=blue,
    urlcolor=blue
}

\begin{document}

\title{SATA: Safe and Adaptive Torque-Based Locomotion Policies Inspired by Animal Learning}

\author{Peizhuo LI$^{1*}$, Hongyi LI$^{1*}$, Ge SUN$^{1}$, Jin CHENG$^{2}$, Xinrong YANG$^{1}$, Guillaume Bellegarda$^{3}$, Milad Shafiee$^{3}$, \\ Yuhong CAO$^{1\dagger}$, Auke Ijspeert$^{3}$, Guillaume SARTORETTI$^{1}$ \\ $^{*}$Equal contribution, $^{\dagger}$Corresponding author, \\ $^{1}$MARMot Lab, National University of Singapore $^{2}$Computational Robotics Lab, ETH Zurich $^{3}$BioRob Lab, EPFL \\Video: \url{https://youtu.be/b1cpTq0Rc5w} Code:\url{https://github.com/marmotlab/SATA}}

\maketitle
\begin{abstract}
Despite recent advances in learning-based controllers for legged robots, deployments in human-centric environments remain limited by safety concerns. Most of these approaches use position-based control, where policies output target joint angles that must be processed by a low-level controller (e.g., PD or impedance controllers) to compute joint torques. Although impressive results have been achieved in controlled real-world scenarios, these methods often struggle with compliance and adaptability when encountering environments or disturbances unseen during training, potentially resulting in extreme or unsafe behaviors. Inspired by how animals achieve smooth and adaptive movements by controlling muscle extension and contraction, torque-based policies offer a promising alternative by enabling precise and direct control of the actuators in torque space. In principle, this approach facilitates more effective interactions with the environment, resulting in safer and more adaptable behaviors. However, challenges such as a highly nonlinear state space and inefficient exploration during training have hindered their broader adoption. To address these limitations, we propose Safe and Adaptive Torque-based locomotion policies inspired by Animal learning(SATA), a bio-inspired framework that mimics key biomechanical principles and adaptive learning mechanisms observed in animal locomotion. Our approach effectively addresses the inherent challenges of learning torque-based policies by significantly improving early-stage exploration, leading to high-performance final policies. Remarkably, our method achieves zero-shot sim-to-real transfer, eliminating the need for additional fine-tuning on hardware. Our experimental results indicate that SATA demonstrates remarkable compliance and safety, even in challenging environments such as soft/slippery terrain or narrow passages, and under significant external disturbances (e.g., pushing/pulling/pressing on the robot, or manually moving individual legs). These results highlight its potential for practical deployments in human-centric and safety-critical scenarios.
\end{abstract}

\IEEEpeerreviewmaketitle

\begin{figure}
    \vspace{0.2cm}
    \centering
    \includegraphics[width=\linewidth,height=6.6cm]{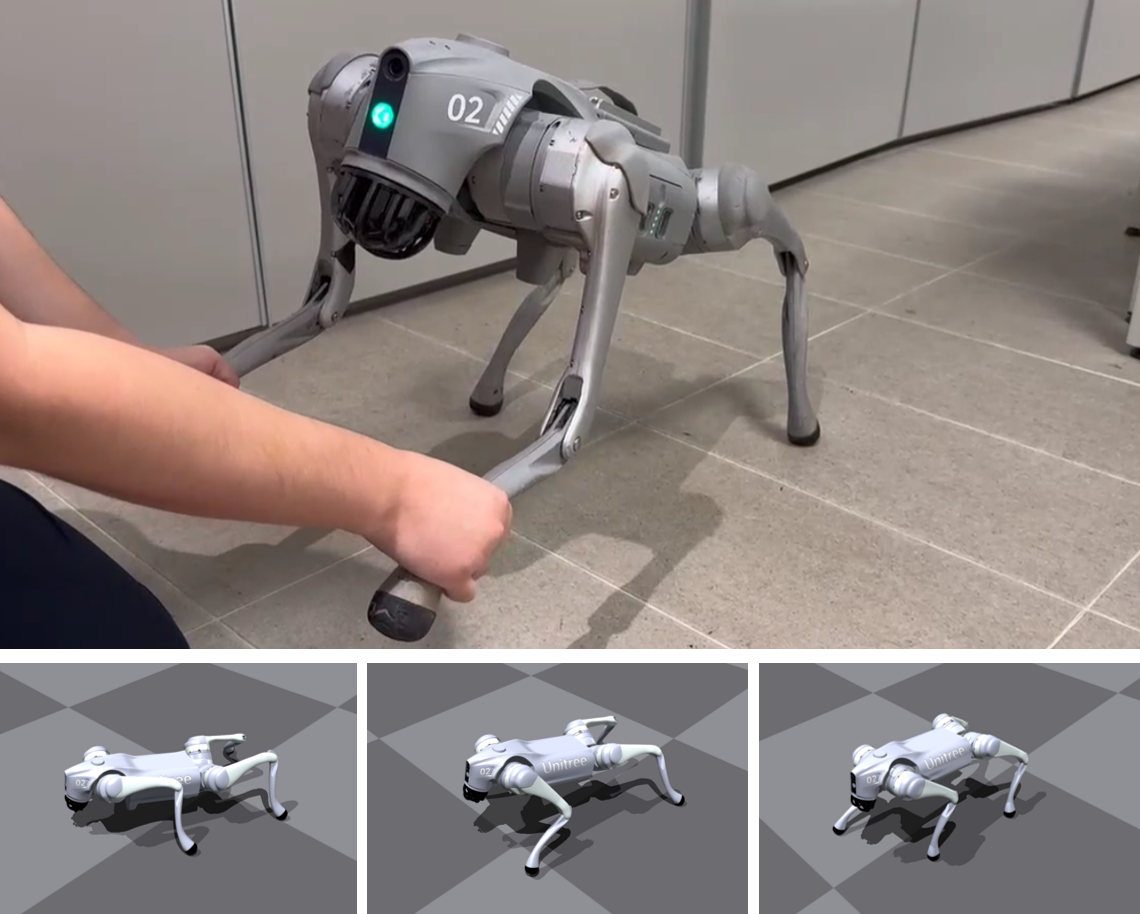}
    \vspace{-0.6cm}
    \caption{Inspired by biomechanical principles and the growth mechanisms of animals in nature, we propose a framework that addresses the challenges of torque-based locomotion learning, achieving zero-shot sim-to-real transfer along with exceptional compliance and safety in challenging environments.}
    \label{fig:enter-label}
    \vspace{-0.6cm}
\end{figure}

\section{Introduction}
\label{sec:introduction}
Reinforcement learning (RL) has demonstrated significant potential in the control of legged robots~\cite{luo2020carl, tan2018sim}. Compared to conventional control methods, such as Model Predictive Control (MPC), RL-based approaches exhibit remarkable robustness, enabling quadrupedal robots to navigate effectively in complex terrain~\cite{choi2023learning, lee2020learning, miki2022learning}.

Most existing RL-based methods rely on position control~\cite{hu2019learning, tsounis2020deepgait}. In this approach, neural networks output target joint positions, which are subsequently translated into joint torque commands through a low-level proportional-derivative (PD) controller. Position-based policies are simple and easy to train, as they abstract away the complexities of actuation physics and dynamics.
However, this simplicity limits the policy's capacity to explore fine-grained and dynamic behaviors, thereby reducing its adaptability and generalization to unseen challenges in real-world environments.
For instance, position-based policies trained on rigid terrains in simulation often struggle to generalize to deformable environments in the real world due to their out-of-distribution nature.
As a result, these methods usually rely heavily on terrain randomization, where the agent trains in large amounts of potential environments to maximize the chances that the final policy may handle real-world environments~\cite{kim2021quadruped, margolis2023walk}.
Furthermore, position-based strategies exhibit notable limitations in compliance.
Lacking the ability to directly regulate joint torques, these methods often lead to overreactions to external disturbances.
For example, if one of the robot's legs suddenly gets stuck, a position-based policy may aggressively attempt to force the actuator into the commanded position. This can cause the motors to generate excessively high torques, potentially destabilizing or even damaging the robot, or creating safety hazards to its surroundings, such as nearby objects or even humans.
 
In contrast, torque-based learning controllers aim to train a policy network that directly generates torques for all joints, providing enhanced compliance and adaptability.
By directly controlling actuation in torque space, this approach enables finer interaction with the environment, leading to more dynamic and robust locomotion.
Moreover, torque control allows the robot to explicitly reason about its dynamics, theoretically offering a greater ability to handle unknown environments and abrupt disturbances without relying on predefined low-level controllers~\cite{sombolestan2024adaptive}.
However, torque-based policies come with their own set of challenges, including a high-dimensional action space and greater nonlinear transformations from states to actions. These factors make exploration substantially harder during training, particularly in the early stages.
During this phase, the abundance of local optima can obstruct the exploration process, leading to premature convergence and unnatural gait behaviors.
Few works have effectively addressed these challenges. DeCAP, proposed by \citet{sood2024decap}, mitigates these issues by leveraging pre-trained position-based policies. However, its increased training complexity limits its broader adoption. Similarly, \citet{chen2023learning} successfully trained a torque policy by incorporating additional reward terms and action scaling. 
Yet, this approach remains highly sensitive to hyperparameter tuning and often exhibits low exploration efficiency during the initial stages of training, preventing it from consistently yielding high-performance policies.

To overcome these limitations, we propose enhancements to directly learn torque-based policies by drawing inspiration from biological systems.
In animals, smooth motion actuation is achieved through the intrinsic biomechanical properties of muscles, such as the force-velocity relationship described by the Hill model~\cite{seow2013hill}, which regulates movement commands and prevents excessive behaviors.
Moreover, certain muscle mechanisms provide feedback that can aid decision-making, such as muscle fatigue (Feedback of pain due to prolonged exertion)~\cite{enoka2008muscle}.
Borrowing from these mechanisms, we integrate a simplified biomechanical model into our torque-based learning framework. This model retains the functional characteristics of biological muscles, enabling the robot to perform smoother actions while mitigating the risk of suboptimal convergence.
In addition, we incorporate a growth mechanism inspired by the gradual development processes observed when animals learn to locomote.
This mechanism dynamically adjusts key robot properties during training, such as torque limits and control frequency, while also modulating the reward terms throughout the process. This significantly enhances early-stage exploration and improves the generalizability of the trained policy.

By addressing the inherent challenges in torque-based policy learning, our approach not only provides a robust and efficient solution for torque-based control but also demonstrates high performance in compliance and adaptability to previously unseen scenarios, such as locomotion on deformable terrains.
These results highlight the potential of torque-based controllers to surpass the limitations of position-based methods, enabling safe and robust locomotion in complex environments.

The main contributions of this work are as follows:

\begin{itemize}
    \item \textbf{Stable and Efficient Torque-Based Learning:}  
    We propose a novel framework for learning torque-based locomotion policies with a growth mechanism that gradually unlocks torque limits, control frequency, and reward terms, enhancing sample efficiency and training stability in torque space.

    \item \textbf{Biomechanical Model and Safety:}  
    We implement a simplified biomechanical model for actuators that ensures smooth and safe behaviors, even under disturbances and performs robustly in diverse environments.

    \item \textbf{Generalization:}  
    Our experimental results demonstrate that SATA generalizes effectively to out-of-distribution terrains and commands while exhibiting exceptional compliance during human interactions. These findings underscore the robustness and versatility of our approach.
\end{itemize}

\section{Related Work}
\label{sec:related_work}

\begin{figure*}
    \centering
    \includegraphics[width=\linewidth,trim=50 150 50 150, clip]{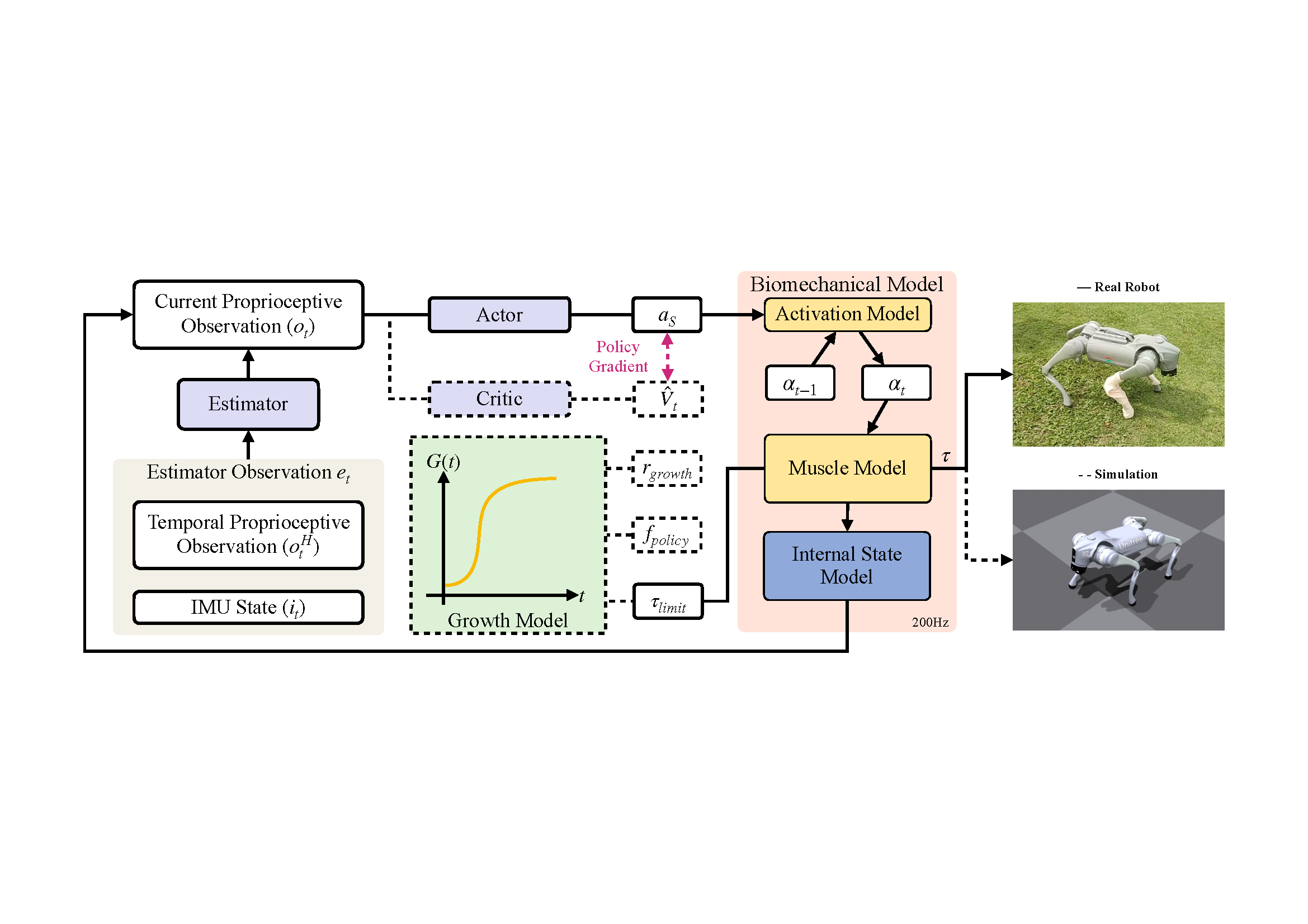}
    \vspace{-1.3cm}
    \caption{Overview of our SATA Framework. Dotted lines indicate parts used only during training, while solid lines indicate those used during both training and deployment. Our framework is mainly composed of 1) a Biomechanical Model (Orange) to ensure the generation of smooth, practical actuator commands \(\tau\) while informing the policy of the current actuator state, and 2) a Growth Model (Green) to help the neural network train a more robust and generalizable policy by gradually adapting rewards \(r_\textit{growth}\), control frequency \(f_\textit{policy}\), and torque limits \(\tau_\textit{limit}\) during training. Finally, we train a state estimator for real world deployment using simulated IMU data and temporal proprioception observations (Grey), to help condition our policy on the (estimated) current robot velocity.}
    \vspace{-0.5cm}
    \label{fig:framework}
\end{figure*}

Quadruped controllers have greatly benefited from the development of deep reinforcement learning (DRL) in recent years, allowing agents to learn impressive controllers~\cite{hu2019learning, bellegarda2022robust, bellegarda2024robust, ren2021learning} that would otherwise require the design and solving of non-linear optimization problems, which often involves approximations that cannot be neglected in real-world settings~\cite{ding2019real, kim2019highly, neunert2018whole, grandia2023perceptive}.
However, the problem of sampling efficiency still hinders its application.
Early approaches employed by~\citet{peng2018deepmimic} combined imitation learning with DRL to solve this problem, demonstrating gaits similar to the training set while adapting to different terrain or disturbances.
This approach allows for the training of adaptive quadruped controllers in highly dynamic acrobatic tasks, but its performance is limited by the dataset quality and robot deployment remains nontrivial.
Addressing these problems, Hwangbo et al.~\cite{hwangbo2019learning} proposed a method to train deployable policies in simulation by introducing an \textit{actuator network} and domain randomization.
Their learned policies can yield different gaits or recover from fallen positions.
To help the robot travel in challenging terrains,~\citet{lee2020learning} introduced a teacher-student framework for quadrupedal robots, allowing them to traverse complex terrains without any visual feedback.
To aid locomotion with environmental information,~\cite{miki2022learning} built upon~\cite{lee2020learning} by integrating additional sensors and employing an attention-based recurrent encoder to fuse proprioceptive and exteroceptive inputs. This resulted in a robust and fast legged motion controller for navigating challenging terrains.
While exteroceptive inputs can enable more informed decision-making, a highly compliant and robust base policy - capable of operating effectively without vision - remains crucial for reliable real-world deployment. Moreover, reliance on exteroception introduces additional challenges, such as the sim-to-real gap, where sensor noise, latency, and real-world variations degrade performance. 

Learning-based controllers typically use position-based action spaces, where the policy directly outputs position commands for the actuators.
These commands are subsequently converted to torque using a low-level (e.g., PD)  controller during training~\cite{aractingi2023controlling, yang2020data, allshire2021laser}.
While such low-level controllers facilitate early-stage exploration for reinforcement learning policies, extensive parameter tuning is often required to ensure successful deployment~\cite{kim2023automated, lyu2024rl2ac}.
Moreover, position control tends to treat the robot as a rigid system, which can result in significant joint or structural stress when operating in uncertain environments (e.g. slipping, collisions), thereby increasing the risk of system damage~\cite{buchli2009compliant}. 

In contrast, torque-based policies, where the policy directly outputs motor torques, eliminate the need for tuning low-level controller parameters.
They also take advantage of the inherent compliance of torque control to reduce structural stress and impact forces, enhancing the safety of human-robot interactions~\cite{calanca2015review}.
There, Chen et al.\cite{chen2023learning} achieved the first successful sim-to-real transfer of end-to-end torque control for quadrupedal locomotion, enabling RL policies to directly predict joint torques at high frequencies.
However, Kim et al.\cite{kim2023torque} highlighted that torque-based state spaces exhibit significant non-linearity and the controllers need to operate at much higher frequencies, which impairs exploration efficiency during early-stage training.

With the advancement of parallel simulation techniques and high-performance computing~\cite{makoviychuk2021isaac, rudin2022learning}, massively parallel RL environments have significantly improved sampling efficiency, reducing the training duration to mere minutes~\cite{rudin2022learningwalkminutesusing}.
This allows training of more complicated frameworks that can be deployed on-robot easily~\cite{zuo2024learning, chen2024learning, kumar2021rma}.
To enhance performance, researchers have explored the integration of learning-based approaches with traditional control techniques. For instance, \citet{gangapurwala2022rloc} proposed RL2AC, which combines reinforcement learning policy with an adaptive torque compensator to mitigate external disturbances and model mismatches caused by the sim-to-real gap.
Similar studies have also demonstrated that integrating the adaptability of learning-based methods with the robustness of traditional models can significantly improve the locomotion performance of quadruped robots in complex environments \cite{yao2021hierarchical, zhang2022learning, zhang2022model}.

The combination of bio-inspired control and reinforcement learning is another major direction to address the adaptability challenges of legged robots in dynamic and complex environments~\cite{abdi2019reinforcement, ouyang2021adaptive}.
This hybrid approach integrates structured prior knowledge from bio-inspired control within reinforcement learning to optimize high-level strategies and key parameters, enabling efficient robot control~\cite{humphreys2024learning, wang2021cpg, wei2018bio, zhang2023synloco}. For instance, Margolis et al.\cite{margolis2023walk} achieved agile locomotion, including walking, trotting, and pronking, by mimicking natural gait patterns through carefully designed rewards.
Similarly, \citet{peng2020learning} employed imitation learning to teach legged robots agile skills by mimicking real-world animals, while \citet{fu2021minimizing} leveraged deep reinforcement learning to minimize energy consumption and achieve emergent gaits.
Inspired by Central Pattern Generators (CPGs) observed in animals, Bellegarda et al.\cite{bellegarda2022cpg} combined CPGs with a DRL framework to generate robust and omnidirectional locomotion.
Building on this foundation, \citet{sun2024learning} introduced the learning-based hierarchical control framework, utilizing a spinal policy to adjust CPG frequency and amplitude for rhythmic movement, and a descending modulation policy to adaptively modify rhythmic outputs for precise control on challenging terrains.
Similar to this idea, various hierarchical methods have also been proposed for legged controllers~\cite{he2024learning, jain2019hierarchical, jain2020pixels}. 
Although these studies have achieved promising results, most of them focus on mimicking animal behaviors or neural structures. Relatively limited research has explored the effects of biomechanical properties and growth processes on the development of locomotion skills in animals, let alone incorporating them into the learning framework for legged robots.

\section{Bio-Inspired Neural Architecture}
\label{sec:control_framework}
To achieve robust and adaptive locomotion control in legged robots, we propose a bio-inspired neural architecture that emulates key principles of biological systems. This architecture comprises two core modules: the Neural Networks and the Biomechanical Model, as illustrated in Fig.~\ref{fig:framework}. While the Neural Networks generate action signals based on proprioceptive information (see detail in section \ref{sec:Neural_network}), the Biomechanical Model is designed to process these action signals, introducing activation dynamics and muscle-like force modulation to ensure smooth and realistic control signals. Additionally, the Biomechanical Model provides internal states feedback to the Neural Networks, providing more temporal information and improved overall performance.

\subsection{Biomechanical Model}
\label{sec:biomechanical_model}
Compared to directly using the neural network's output as joint torques, our approach aims to reduce exploration difficulty during training and improve motion continuity. To achieve this, we refine the action signal \(a_s\) generated by the neural network using a biomechanical model that employs a biologically inspired two-step process and incorporates a feedback mechanism, including a fatigue mechanism. This fatigue mechanism, inspired by biological systems, dynamically quantifies actuator load and recovery, contributing to more balanced and robust control. The biomechanical model ensures biologically plausible and stable locomotion through three key components.

\begin{itemize}
    \item \textbf{Activation Model}: Converts action signals into intermediate activation signals, incorporating temporal smoothing to reflect the sequential nature of biological systems. This process ensures continuity and prepares signals for precise torque generation.
    
    \item \textbf{Muscle Model}: Transforms activation signals into joint torques by loosely approximating muscle dynamics. This approach limits torque output to a safe range, preventing abrupt changes that could destabilize the system.
    
    \item \textbf{Internal State Model}: Tracks the fatigue state of actuators, providing real-time feedback to the neural network. This feedback helps to optimize load distribution, enhancing stability during training and deployment.
\end{itemize}

By combining these components, the biomechanical model improves exploration efficiency, reduces the risk of local optima, and bridges the sim-to-real transfer gap, making torque commands practical and effective for real-world applications.

\subsubsection{Activation Model}

Output by our policy network, the action signal \(a_s\) first passes through the activation model~\cite{botterman1986gradation}. This model functions similarly to motor neurons~\cite{caillet2023motoneuron, callahan2013computational}, transforming the action signal into the corresponding activation signal, \(\alpha_{\text{current}}\):
\begin{equation}
    \alpha_{\text{current}} = \tanh\left(\frac{a_s \cdot \kappa_{\text{scale}}}{\tau_{\text{limit}}}\right).
    \label{eq:rescale}
\end{equation}
Here, \(\kappa_{\text{scale}}\) represents a scaling factor analogous to the gain of motor neurons in translating neural commands into muscle activations, and \(\tau_{\text{limit}}\) denotes the torque limits of each motor. The resulting activation signal, constrained within the range \((-1, 1)\), emulates the antagonistic coordination of joint control by opposing muscle groups. For example, \(1\) represents full contraction of one muscle and complete relaxation of its antagonist, resulting in a clockwise torque, while \(-1\) is the exact opposite, enabling precise joint torque modulation.

To ensure smooth and continuous activation dynamics, this model incorporates a temporal update mechanism inspired by the "Hysteresis Effect"~\cite{hassani2014survey}. This mechanism accounts for the influence of prior activations on the current state, ensuring stability and natural transitions in movement. The simplified first order temporal evolution of the activation signal \(\alpha_t\) is governed by:
\begin{equation}
    \alpha_t = (\alpha_{\text{current}} \cdot \gamma) + (\alpha_{t-1} \cdot (1 - \gamma)).
    \label{eq:activation_process}
\end{equation}
In this formulation, \(\gamma\) serves as a smoothing factor.  This ensures a smooth temporal evolution of activation signals, which promotes continuity and smoothness in movements. 

\subsubsection{Muscle Model}
The activation signal is subsequently passed to the muscle model, where it is transformed into joint torques. There, we developed a dynamics model loosely inspired by the classical Hill muscle model to mitigate potential extreme behaviors~\cite{schmitt2019dynamics}. Specifically, we focused on modeling and optimizing the force-velocity relationship in the Hill model~\cite{ romero2016comparison, till2008characterization}. When the activation signal aligns with the joint motion direction, the torque output decreases as the joint velocity increases, effectively suppressing rapid and extreme torque signals. Conversely, when the joint velocity opposes the activation signal, the model generates greater torque to drive the joint towards the desired motion velocity. We believe this force-velocity relationship enhances the activation network's sensitivity to dynamic motion states, reducing instability risks during training and improving exploration efficiency.

Specifically, the joint torque \(\tau\) is computed as:
\begin{equation}
    \tau = \tau_{\text{limit}} \cdot \alpha_t \cdot \left(1 - \text{sign}(\alpha_t) \cdot \frac{\dot{q}}{\dot{q}_{\text{limit}}}\right),
    \label{eq:hill_model}
\end{equation}
\noindent where \(\dot{q}\) represents the joint velocity, and \(\dot{q}_{\text{limit}}\) is the maximum velocity of each joint. 
This model integrates the activation signal with the dynamic characteristics of joint motion, enabling accurate modeling of joint behavior and ultimately improving the stability and efficiency of the overall control system.

\subsubsection{Internal State Model}  
Our internal state model does not directly participate in torque calculation but serves as a feedback mechanism to provide the robot with more information. In this module, we simulate the fatigue mechanism observed in animals to construct a dynamic state indicator that quantifies the accumulation and recovery of fatigue during the robot's operation~\cite{boonstra2008fatigue, cowley2017inter}. The fatigue indicator is directly influenced by the joint torque intensity and evolves dynamically over time, following the equation:

\begin{equation}
    \zeta_t = (\zeta_{t-1} + |\tau| \cdot dt) \cdot \beta,
    \label{eq:fatigue}
\end{equation}
where \(\zeta\) represents the fatigue state, \(dt\) the time step, and \(\beta\) the recovery factor, describing the rate at which fatigue dissipates over time.

Through this mechanism, the robot can continuously update its fatigue state, providing the neural network with a dynamic feedback signal. 
This fatigue feedback effectively optimizes the control strategy, preventing certain actuators from operating under prolonged high loads, which could otherwise lead to excessive wear or reduced performance. 
Furthermore, it enables the robot to distribute motion loads more efficiently on each leg during task execution, reducing the occurrence of suboptimal behaviors, such as over-reliance on three-legged or two-legged motion patterns during exploration. 

\subsection{Neural Networks}
\label{sec:Neural_network}
Our SATA framework comprises three sub-networks: actor, critic, and estimator. The actor and critic are optimized using the Proximal Policy Optimization (PPO) algorithm, while the estimator is trained separately, using supervised learning.

\subsubsection{Observation and Action Space for Actor}
The proprioceptive observation vector \(\boldsymbol{o_t} \in \mathbb{R}^{60}\) serves as the input to the actor network, encapsulating comprehensive information about the robot's state.

The observation vector is structured as:
\[
o_t = [v_t, w_t, g_t, q, \dot{q}, v_{\text{cmd}}, \tau, \zeta]^T,
\]
where \(v_t \in \mathbb{R}^3\) represents the linear velocity of the robot base. During training, \(v_t\) is directly sourced from the simulator, while during deployment, it is estimated by the estimator network. The estimator uses historical proprioceptive and inertial data, structured as:
\[
e_t = [[o_{t-10}^H, i_{t-10}], ..., [o_t^H, i_t]]^T.
\]
Here, \(o_t^H = [q, \dot{q}]\) includes the joint angles and velocities, and \(i_t = [\mathbf{a}_t, \boldsymbol{\omega}_t, \mathbf{g}_t]\) represents the IMU data---featuring linear acceleration \(\mathbf{a}_t \in \mathbb{R}^3\), angular velocity \(\boldsymbol{\omega}_t \in \mathbb{R}^3\), and gravity direction \(\mathbf{g}_t \in \mathbb{R}^3\). This design enables robust velocity estimation during deployment.

Other components of \(o_t\) include \(w_t \in \mathbb{R}^3\), the angular velocity of the robot base, and \(g_t \in \mathbb{R}^3\), the gravity direction vector in the body frame. These quantities aid in orientation estimation and maintaining balance. Additionally, \(v_{\text{cmd}} \in \mathbb{R}^3\) specifies the desired linear and angular velocities, while \(q \in \mathbb{R}^{12}\), \(\dot{q} \in \mathbb{R}^{12}\), \(\tau \in \mathbb{R}^{12}\), and \(\zeta \in \mathbb{R}^{12}\) represent the joint states, joint velocities, joint torques, and fatigue state, as described earlier.

Based on \(o_t\), the actor policy generates the action \(a_s \in \mathbb{R}^{12}\), which represents the desired joint torques. These torques are further refined by the biomechanical model to ensure smooth and stable control.

\subsubsection{Reward Design}
In this work, we adopt a relatively simple reward structure, made up of 9 terms designed to effectively encourage natural and robust locomotion. These rewards are categorized into two types: locomotion objectives and auxiliary posture maintenance rewards. The details are summarized in Table~\ref{tab:rewards}.

\begin{table}[h!]
\centering
\renewcommand{\arraystretch}{1.3}
\setlength{\tabcolsep}{8pt}
\caption{Reward components and weights (\(dt=0.005\))}
\label{tab:rewards}
\vspace{-0.3cm}
\begin{tabular}{lcc}
\toprule
\textbf{Reward Terms} & \textbf{Equation} & \textbf{Weight} \\ 
\midrule
\multicolumn{3}{l}{\textbf{Locomotion Objectives}} \\ 
\(r_{\text{tracking},x}\) & \(\phi\left(v_x - v_x^{\text{cmd}}\right)\) & \(10dt\) \\ 
\(r_{\text{tracking},y}\) & \(\phi\left(v_y - v_y^{\text{cmd}}\right)\) & \(5dt\) \\ 
\(r_{\text{tracking},\text{yaw}}\) & \(\phi\left(\omega_{\text{yaw}} - \omega_{\text{yaw}}^{\text{cmd}}\right)\) & \(5dt\) \\ 
\midrule
\multicolumn{3}{l}{\textbf{Auxiliary Posture Maintenance}} \\ 
\(r_{\text{base height}}\) & \(\min(h_b, h_t)\) & \(5dt\) \\ 
\(r_{\text{roll}}\) & \(|g_{\mathrm{y}}|\) & \(-5dt\) \\ 
\(r_{\text{velocity},z}\) & \((v_z)^2\) & \(-5dt\) \\ 
\(r_{\text{joint limits}}\) & $\sum \left[ (q_{\min} - q)^+ + (q - q_{\max})^+ \right]$ & \(-5dt\) \\ 
\(r_{\text{fatigue}}\) & \(\zeta \cdot |\tau_{d} \cdot \kappa_{\text{scale}}|\) & \(-0.05dt\) \\ 
\(r_{\text{joint acceleration}}\) & \(\ddot{q}^2\) & \(-1e-6dt\) \\ 
\bottomrule
\end{tabular}
\vspace{-0.5cm}
\end{table}

There, \(\phi(x) = e^{-4 \cdot |x|}\) represents a Gaussian-shaped function used to penalize deviations between actual and commanded values. \(h_b\) and \(h_t\) denote the robot's base height and target base height above the ground, respectively. \(q_{\text{min}}\) and \(q_{\text{max}}\) define the lower and upper limits of each joint, while \(\ddot{q}\) represents the joint acceleration.

\section{Growth-Based Training}
\label{sec:growth}
Due to the highly nonlinear nature of the torque space, training a torque-based policy poses greater challenges than a position-based one, especially during early-stage exploration. To address this, we propose a biologically inspired growth mechanism that mimics animal development by progressively unlocking the robot's physical capabilities, dynamically adapting reward functions, and gradually increasing control frequency. This process facilitates more efficient policy learning while preserving stability and promoting generalization.

While related in spirit to curriculum learning~\cite{qin2019sim,yu2018learning,li2024learning,kobayashi2020reinforcement} and progressive learning~\cite{berseth2018progressive,li2023robust} approaches - which typically increase task difficulty over time - our method maintains a fixed task throughout training. Instead of staging increasingly complex goals, we focus on enhancing the agent's embodiment by gradually expanding what it is physically allowed to do. This leads to deeper exploration and reduces the risk of suboptimal shortcuts, such as exploiting a single powerful joint. Our strategy also aligns conceptually with reward scheduling techniques, where the learning signal evolves in tandem with the agent's growing capabilities~\cite{riedmiller2018learning,faust2019evolving}. For instance, as the agent develops from basic contact with the ground to full stepping behavior, the reward emphasis naturally shifts to reflect the current developmental stage. 

\subsection{Implementation of the Growth Mechanism}
Instead of granting the policy full access to the action space from the start of training, we propose that partially limiting the robot's abilities can promote more efficient exploration. 
Additionally, gradually increasing the control frequency simplifies exploration and mitigates the problem of delayed reward during the early stages of training.
To unify these components, we introduce a time-dependent general scale \( G(t) \), derived from the Gompertz model~\cite{ardelean2005estimation}, a well-established framework for modeling growth:
\begin{equation}
G(t) = e^{-e^{-k \cdot (t - t_{0})}}.
\label{eq:general_scale}
\end{equation}
Here, \( G(t) \) serves as the basis for dynamically adjusting training parameters. The parameters \( k \), \( t \), and \( t_{0} \) denote the growth rate, the current training step, and the step at which the maximum growth rate occurs, respectively.

\subsubsection{Torque Limits and Control Frequency Adjustment}

Using \( G(t) \), we dynamically update the torque limits (\( \tau_{\text{limit}} \)) and control frequency (\( f_{policy} \)), which are closely linked to growth~\cite{larsson1979muscle, gage2004structural, stampanoni2019synaptic}, during training. These updates enable the robot to gradually get access to its full operational capabilities:
\begin{equation}
    \tau_{\text{limit}} = \tau_{\text{start}} + (\tau_{\text{end}} - \tau_{\text{start}}) \cdot G(t),
    \vspace{-0.2cm}
    \label{eq:update_tau}
\end{equation}
\begin{equation}
    f_{policy} = f_{\text{start}} + (f_{\text{end}} - f_{\text{start}}) \cdot G(t).
    \label{eq:update_freq}
\end{equation}
Here, \(\tau_{\text{start}}\) and \(f_{\text{start}}\) represent the initial torque limit and control frequency at the beginning of training, while \(\tau_{\text{end}}\) and \(f_{\text{end}}\) denote their maximum values, which are reached as training progresses.

\subsubsection{Dynamic Adjustment of Reward Expressions}
We also leverage \( G(t) \) to dynamically adjust certain reward expressions during training. This approach mirrors how animals maintain a consistent overall goal while shifting their focus across different learning stages. For instance, animals prioritize balance and upright posture during early locomotion learning, then focus on stepping, and eventually smoothen their movements. Similarly, \( G(t) \) allows the robot to adapt reward priorities to align with specific training objectives. The adjusted growth-based reward \(r_{growth}\) expressions are summarized in Table~\ref{tab:adjusted_rewards}:

\begin{table}[h]
\Large
\centering
\renewcommand{\arraystretch}{1.3} 
\setlength{\tabcolsep}{1pt}
\caption{Details of our Growth-based reward terms}
\vspace{-0.3cm}
\resizebox{0.9\columnwidth}{!}{
\begin{tabular}{cc}
\hline
\textbf{Adjusted Rewards} & \textbf{Calculation} \\ \hline
$r_{\text{tracking}, x}$ & $\phi\left(v_{x} - \frac{v^{\text{cmd}}_{\text{x,max}} + v^{\text{cmd}}_{\text{x,min}}}{2}\right) (1 - G(t)) + \phi\left(v_{x} - v_{x}^{\text{cmd}}\right) (1 + G(t))$ \\[1ex]
$r_{\text{tracking}, y}$ & $\phi\left(v_{y} - v_{y}^{\text{cmd}}\right) G(t)$ \\[1ex]
$r_{\text{tracking}, \text{yaw}}$ & $\phi\left(\omega_{\text{yaw}} - \omega_{\text{yaw}}^{\text{cmd}}\right) G(t)$ \\[1ex]
$r_{\text{base height}}$ & $\min(h_{\text{b}}, h_{\text{t}}) (1 + G(t)) -\max\left(g_x, -\min\left(0, 0.2 \cdot (1.5 - 2 \cdot G)\right)\right)$ \\ \hline
\end{tabular}
}
\label{tab:adjusted_rewards}
\end{table}

As \( G(t) \) evolves, the rewards shift focus from encouraging basic behaviors, such as forward motion, to more complex objectives like maintaining body height and tracking precise velocity commands. This progression enables the robot to transition from simple motions to refined locomotion and compliant posture control efficiently.

\subsection{Training Details}

We conduct training using Isaac Gym and the Unitree GO2 quadruped robot. This framework enables high-throughput simulation, allowing us to simulate 4096 instances of the GO2 robot in parallel on a single NVIDIA RTX 4090 GPU. We utilize Proximal Policy Optimization (PPO) to train the control policy. The hyperparameters and neural network architecture are consistent with \cite{rudin2022learning}, including a multilayer perceptron (MLP) with three hidden layers, whose hidden dimensions are \([512, 256, 128]\). Leveraging this framework, we achieve efficient policy learning within 20 minutes/ 3000 episodes.

The maximum episode length is set to 10 seconds. The environment resets when the robot flips over or its joint angles exceed predefined limits. The terrains include rough surfaces (with a maximum height variation of 12 cm) and slopes.

After each reset, the robot is repositioned lying flat on the ground, with varying levels of motor fatigue already applied. This random initialization enhances the generalization capability of the learned policy. Target velocity commands \([v^{\text{cmd}}_x, v^{\text{cmd}}_y, \omega^{\text{cmd}}_{\text{yaw}}]\) are sampled every 5 seconds, with ranges set to \(v^{\text{cmd}}_x \in [-0.5, 1.5]\)~m/s, \(v^{\text{cmd}}_y \in [-0.5, 0.5]\)~m/s, and \(\omega^{\text{cmd}}_{\text{yaw}} \in [-1.5, 1.5]\)~rad/s. 

Domain randomization is applied during training to simulate real-world variability. The specific randomization settings are as follows:
\begin{itemize}
    \item \textbf{Added base mass:} Randomly increased by up to 5 kg.
    \item \textbf{Ground coefficient of friction:} Varied within [0.5, 1.25].
    \item \textbf{Probability to hold last actions or observations:} 10\%.
    \item \textbf{Shifted center of mass:} Varied within [-0.2, 0.2] m along the x-axis, and [-0.1, 0.1] m along the y- and z-axes.
\end{itemize}

During training, control frequency and torque limits are progressively increased, asymptotically approaching their maximum values without fully reaching them, as defined in Eqs.\ref{eq:update_tau} and\ref{eq:update_freq}. For deployment, the robot's full capacity is restored by setting these parameters to their respective maximums,  \(f_{\text{end}}\) and \(\tau_{\text{end}}\). All hyperparameters related to the growth schedule and biomechanical model are summarized in Table~\ref{tab:training_params}.

\begin{table}[h]
\centering
\setlength{\tabcolsep}{5pt} 
\caption{Growth schedule and model hyperparameters}
\label{tab:training_params}
\vspace{-0.3cm}
\resizebox{0.9\columnwidth}{!}{%
\begin{tabular}{|c|c|c|c|c|c|}
\hline
\multicolumn{6}{|c|}{\textbf{Growth Schedule}} \\
\hline
$k$ & 0.00003 & $t_0$ & 24000 & $\tau_{\text{start}}$ & 7.05 Nm \\
\hline
$\tau_{\text{end}}$ & 23.5 Nm & $f_{\text{start}}$ & 100 Hz & $f_{\text{end}}$ & 200 Hz \\
\hline
\multicolumn{6}{|c|}{\textbf{Biomechanical Model}} \\
\hline
$\kappa_{\text{scale}}$ & 5.0 & $\gamma$ & 0.6 & $\beta$ & 0.9 \\
\hline
\end{tabular}
}
\vspace{-0.5cm}
\end{table}

\section{Experiments}
\label{sec:experiment}
\subsection{Simulation Experiments}

\subsubsection{Ablation Study}
\label{sec:exp_sim_ablation}
To evaluate the contribution of each component of our approach, we compare the performance of the complete framework (SATA) with variants that remove the biomechanical model (SATA w/o biomechanical model) or the growth mechanism (SATA w/o Growth).

\begin{figure}[tbp]
    \vspace{-0.8cm}
    \centering
    \includegraphics[width=\linewidth, trim=125 75 125 75, clip]{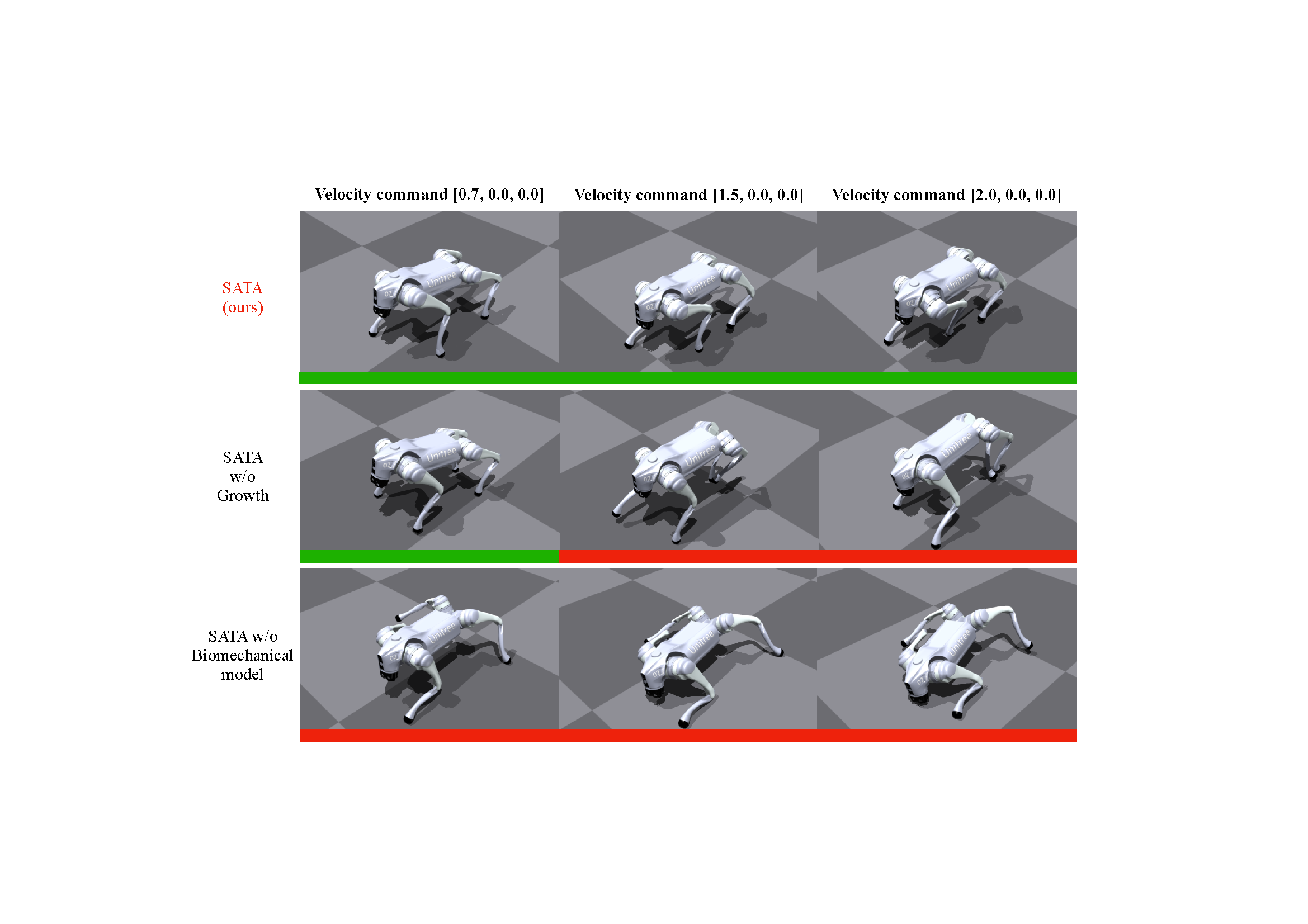}
    \vspace{-1.35cm}
    \caption{Ablation study of the proposed framework, showing successful training in green and failure/premature convergence in red. SATA is compared with variants that lack the biomechanical model or the growth mechanism. Notice that without the growth model (SATA w/o Growth), the policy struggles to achieve high commanded velocities (1.5m/s), especially above the range seen during training. Without the biomechanical model (SATA w/o biomechanical model), the robot is completely unable to learn a coherent gait, instead learning to shift its feet on the floor asymmetrically.}
    \label{fig:ablation_study}
    \vspace{-0.5cm}
\end{figure}

\begin{figure*}
    \centering
    \includegraphics[width=\linewidth, trim=50 100 50 100, clip]{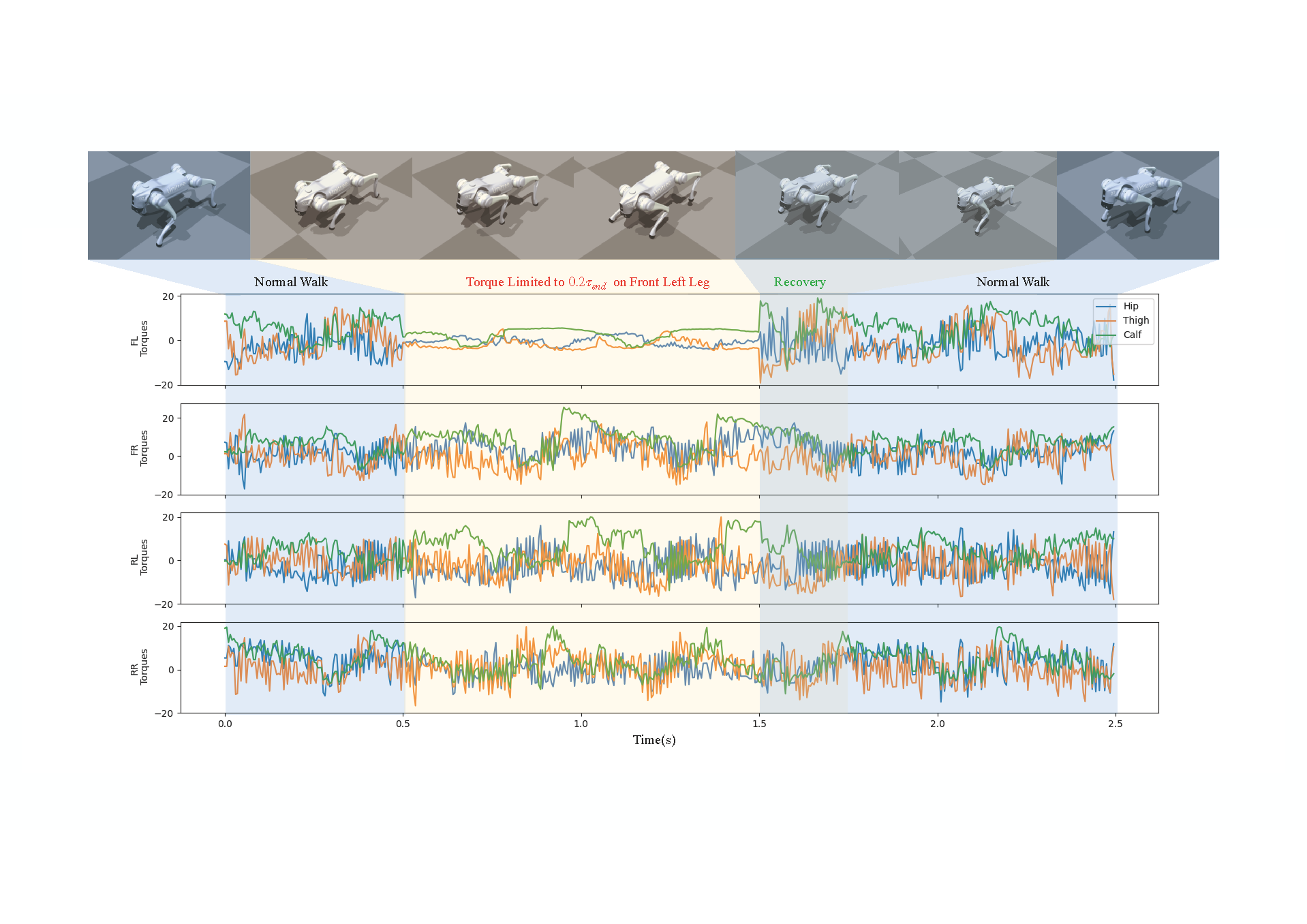}
    \vspace{-1.1cm}
    \caption{Response to a sudden torque limitation on the front left leg (at $t = 0.5\,\mathrm{s}$). During this disturbance ($0.5\,\mathrm{s} < t < 1.5\,\mathrm{s}$), the robot dynamically compensates using other legs, and rapidly recovers once the torque is restored ($1.5\,\mathrm{s} < t < 1.75\,\mathrm{s}$).
}
    \vspace{-0.5cm}
    \label{fig:torque_limitation_single_leg}
\end{figure*}

As shown in Fig.~\ref{fig:ablation_study}, the biomechanical model plays a critical role in enabling natural and stable locomotion. When this biomechanical model is removed, the robot converges to unnatural gaits, such as three-legged support patterns, which reduce stability and limit energy efficiency. This highlights the importance of the biomechanical model and feedback mechanisms in smoothing motion commands and preventing suboptimal convergence.

\begin{figure}[t]
    \centering
    \includegraphics[width=0.9\linewidth, height=2.5cm]{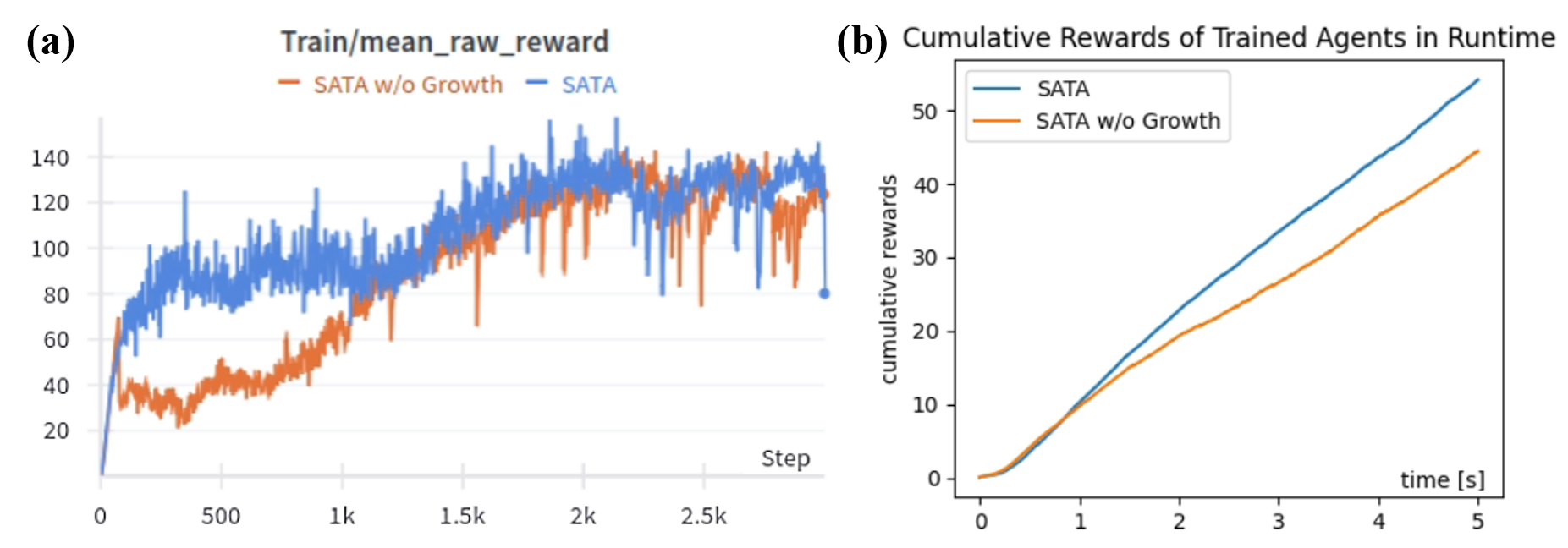}
    \vspace{-0.45cm}
    \caption{Comparison of SATA and SATA w/o Growth. Training rewards (a), without G(t) adaptation, and cumulative rewards in simulation test (b), when commanded to run at 1.8 m/s (slightly OOD).}
    \label{fig:growth-comparison}
    \vspace{-0.5cm}
\end{figure}

On the other hand, the inclusion of the growth mechanism leads to higher early-stage training efficiency and shows better generalization when tracking out-of-distribution (OOD) velocity commands. As demonstrated in Fig.~\ref{fig:growth-comparison}a, SATA significantly outperforms SATA w/o growth in early stages of training, demonstrating the impact of this mechanism in early stage exploration. Moreover, when comparing the cumulative reward of both scenarios under OOD velocity commands ($v_x=1.8m/s$) as in Fig.~\ref{fig:growth-comparison}b, we can see that our method outperforms SATA w/o growth, demonstrating the impact of the growth mechanism on policy generalization. Upon closer inspection, as in Fig.~\ref{fig:ablation_study}, a risky gait emerges as the policy is commanded to the highest command seen in training and in OOD scenarios, leading to unstable tracking of velocity commands.

In particular, these results suggest that the complementary roles of the biomechanical model help ensure stable and natural motion, and the growth mechanism enhance the policy's adaptability and robustness under diverse conditions.

\subsubsection{Robustness to Single-Leg Failure}
In this experiment, we simulate the failure of a single leg by abruptly reducing the maximum torque of its motor to 20\% of its original capacity (\(0.2 \tau_{\text{end}}\)). By doing so, we validate the robustness of our policy  during asymmetric conditions.

As shown in Fig.~\ref{fig:torque_limitation_single_leg}, when the front left leg's torque output is limited, the other legs adaptively increase their force output to stabilize the robot's posture and prevent a collapse. This dynamic redistribution of effort ensures continuous and stable locomotion even under single leg failures. Once the torque capacity of the front left leg is restored, the robot reactively transitions back to its normal walking gait, demonstrating the efficiency of the adaptive feedback mechanism in handling and recovering from localized perturbations.

\begin{table}[ht]
\centering
\caption{Performance Comparison between Our Method and Baselines Across Different Robustness Tests (5 Trials per Test)}
\vspace{-0.3cm}
\label{tab:performance_comparison}
\resizebox{\linewidth}{!}{
\begin{tabular}{lcccc}
\hline
                              & \textbf{\begin{tabular}[c]{@{}c@{}}Sideway \\ Pushing\end{tabular}} & \textbf{\begin{tabular}[c]{@{}c@{}}Soft \\ Terrain\end{tabular}} & \textbf{Tunnel} & \textbf{\begin{tabular}[c]{@{}c@{}}Vertical \\ Stomp\end{tabular}} \\ \hline
\textbf{SATA (Ours)}          & \textbf{100\%}                                                     & \textbf{100\%}                                                   & \textbf{100\%}  & \textbf{100\%}                                                        \\
\textbf{WalkTheseWay}         & 20\%                                                               & 20\%                                                             & 20\%            & 20\%                                                                  \\
\textbf{Vanilla Position-based Policy} & 80\%                                                      & 40\%                                                             & 0\%             & 20\%                                                                  \\
\textbf{DeCAP (Pure Torque)} & 0\% & 0\% & 0\% & 60\% \\
\textbf{DeCAP (Position-assisted)} & 80\% & 40\%  & 0\% & 80\% \\ 
\textbf{Unitree Built-in MPC} & 80\%                                                               & 100\%                                                            & 0\%             & 80\%                                                                  \\ \hline
\end{tabular}
}
\vspace{-0.5cm}
\end{table}

\subsection{Lab Level Experiments}

To validate the effectiveness of our approach, we deployed it on a Unitree Go2 quadruped robot in real-world scenarios. We also compared its performance against several baseline methods, including Unitree's built-in, MPC-based controller, a vanilla learning position-based policy, a well-known learning position-based policy, WalkTheseWay~\cite{margolis2023walk}, as well as a torque-based policy, DeCAP~\cite{sood2024decap}. Experimental results demonstrated that our approach exhibited robust performance across various scenarios, including handling unexpected disturbances and navigating unseen environments not encountered during training. In multiple evaluations, as summarized in Table~\ref{tab:performance_comparison}, our method consistently outperformed the baseline approaches across all robustness tests (please refer to our associated video for details).

Contrary to the common perception that torque-based methods suffer from a significant sim-to-real gap, our approach achieved zero-shot transfer without any fine-tuning and demonstrated highly stable operation over extended deployment periods. To assess its compliance and adaptability, we conducted a series of external disturbance tests in a controlled lab environment to observe and compare the controller's responses against baseline methods. In the first subsection, we illustrate the compliance of our method during \textbf{human-robot interactions}, while Sections~\ref{sec:impact_on_leg} and~\ref{sec:kicking_stomping} highlight its robustness against \textbf{external disturbances}.
\begin{figure}[tbp]
    \centering
    \includegraphics[width=1\linewidth]{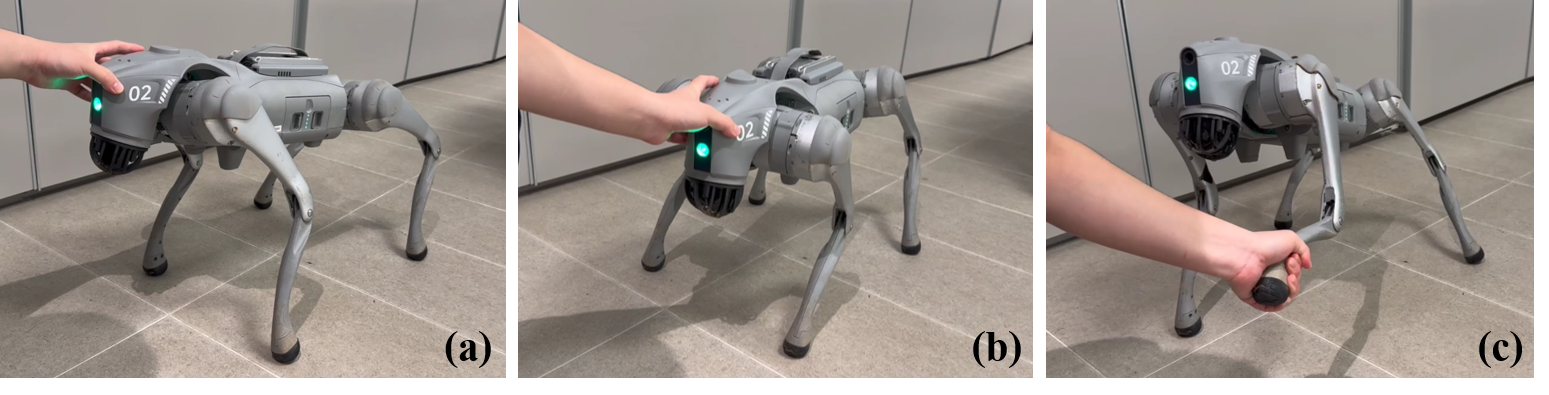}
    \vspace{-0.7cm}
    \caption{Pushing the robot to its left (a) and right (b), and manually lifting its legs (c). In all those cases, the controller did not start trotting nor generate overreacting/hazardous motions.}
    \label{fig:base_range}
    \vspace{-0.5cm}
\end{figure}

\begin{figure}[tbp]
    \centering
    \includegraphics[width=1\linewidth]{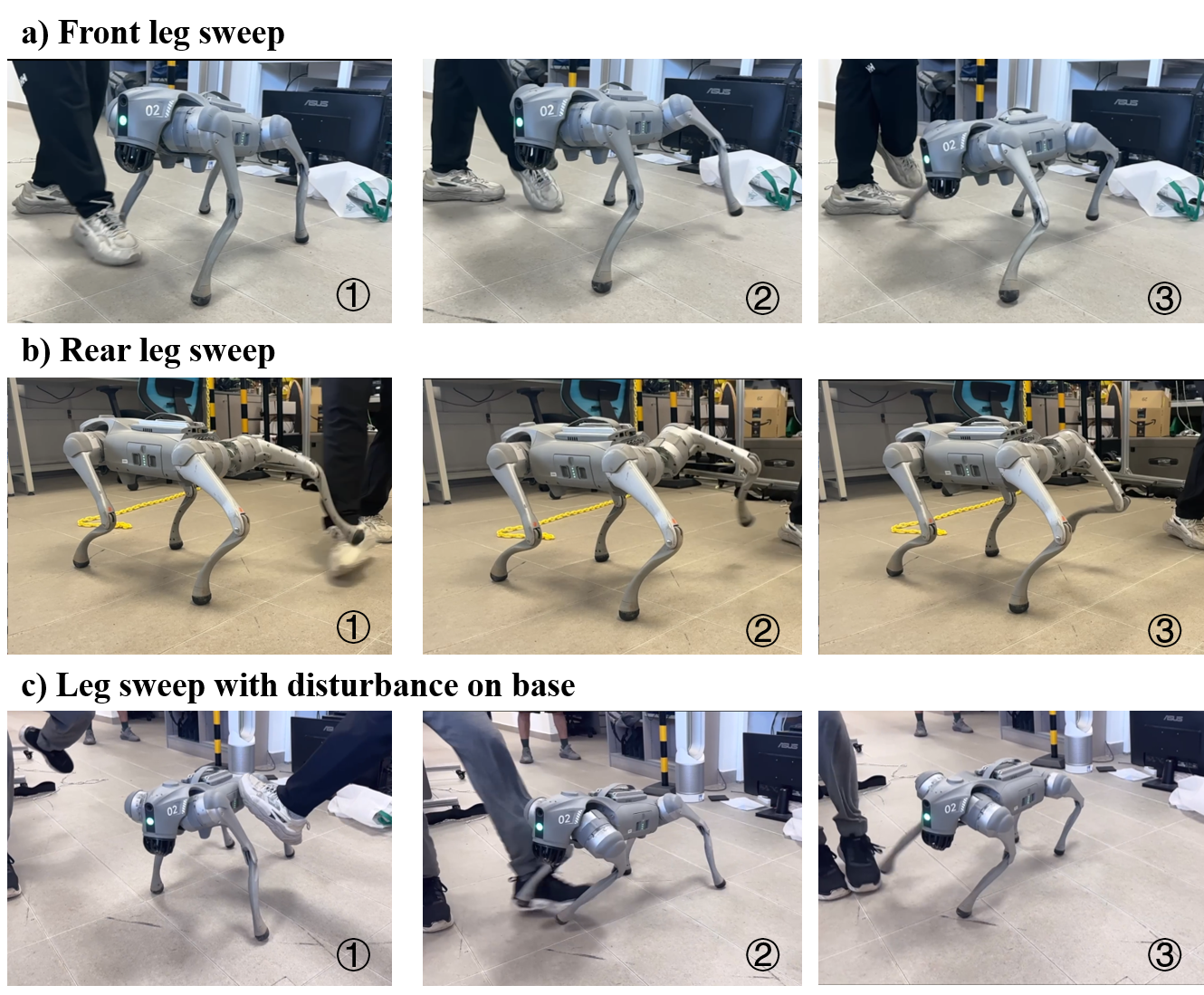}
    \vspace{-0.8cm}
    \caption{Leg disturbance test with a) Backward sweep of the front legs, b) Backward sweep of the back legs, and c) Forward sweep of the front legs right after kicking its body. In a), the rear left leg also lifts up to help balance the whole body, while in b) the robot simply shifts its weight to the other three feet in response to the sweep.}
    \vspace{-0.8cm}
\label{fig:leg_disturbance_test}
\end{figure}

\subsubsection{Safety for Human-Robot Interaction}
\label{sec:compliance}

For robotic manipulators, compliance is critical for ensuring human safety during collaborative tasks. A compliant robotic arm can be easily pushed by a person and will stop applying excessive force upon contact, preventing injuries instead of forcefully moving to a predetermined position. Similarly, compliance in quadrupedal robots is essential for safe interaction with humans. This section focuses on evaluating the passive compliance of our controller and its interaction capabilities with humans.

In standing posture, our torque-based controller allows the robot to respond naturally to human-applied forces, adjusting its body posture without unnecessary stepping motions. The robot only takes corrective steps when balance is at risk. Fig.~\ref{fig:base_range} illustrates the level of compliance the robot can achieve during human robot interaction. These results significantly surpass the performance of position-based controllers, which exhibit a very stiff behavior and are nearly unable to be displaced without stepping. 

\begin{figure}[t]
    \centering
    \includegraphics[width=1\linewidth]{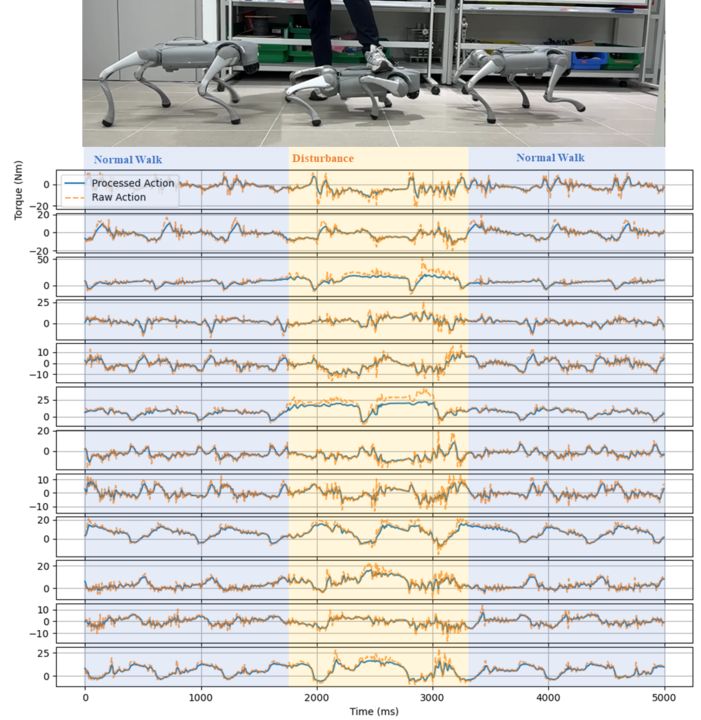}
    \vspace{-0.7cm}
    \caption{Walking under external downward forces/presses. The blue line is the actual torque command given to the motors after the processing done by our biomechanical model, while the orange dotted line is the raw action output of the policy. The robot can continue progressing forward even when an external force press it to the ground. Note that the biomechanical model is crucial in ensuring the safety of the robot especially during these disturbances, limiting excessive torque output and mitigating oscillations.}
    \label{fig:stomp}
    \vspace{-0.5cm}
\end{figure}

\subsubsection{Impact on Legs}
\label{sec:impact_on_leg}

Compliance also plays a crucial role in responding to localized external disturbances. To test this, we placed our robot in a standing posture (\(v^{cmd} = 0\)) while its legs were subjected to forward and lateral sweeps with sufficient force to lift up its foot. As demonstrated in Fig.~\ref{fig:leg_disturbance_test}, the robot's controller exhibited robust performance, successfully resisting these disturbances across all four legs without overreacting. Furthermore, even when additional external impacts were applied to other parts of the body, the controller consistently enabled the robot to regain balance through a series of smooth, adaptive adjustments, quickly regaining balance with a side step.

\subsubsection{Kicking and Stomping}
\label{sec:kicking_stomping}
Most quadrupedal controllers demonstrate robust responses to kicks. However, the impact limit for these controllers varies. Some struggles with recovery from downward forces applied directly from above, while others perform poorly from horizontal impacts that can flip the robot over. In contrast, our controller exhibits remarkable recovery capabilities to these disturbances. Fig.~\ref{fig:stomp} shows how our approach handles a downward stomp during locomotion, with its biomechanical model helping to ensure safe motor commands. 
Furthermore, we note that our method allows the robot to successfully start operating from arbitrary configurations, such as lying flat or standing upright, similar to animals. This capability is rarely observed in position-based methods or traditional control approaches, which typically require the robot to stand up first through predefined instructions before operation.

\subsection{Out-of-distribution Environments}
\begin{figure}[tbp]
    \centering
    \includegraphics[width=1\linewidth]{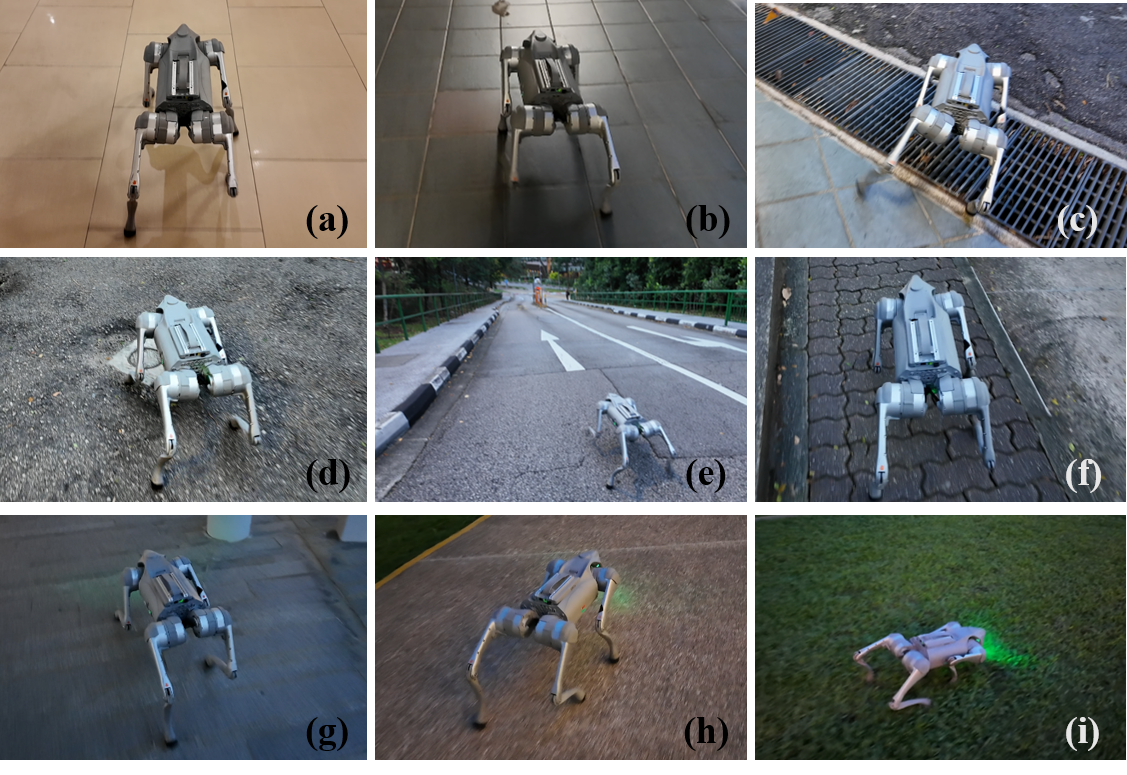}
    \vspace{-0.8cm}
    \caption{Stability test of our proposed controller over a long (1.2km) route, which covers tiled floor indoors (a,b), rough hard unstructured road (c,d), high slopes (e), pedestrian paths (f,g,h), and soft lawn (i). The whole route was completed without human correction, beyond manual adjustment of the robot's heading to chart its course.}
    \label{fig:long_walk}
    \vspace{-0.7cm}
\end{figure}

Fig.~\ref{fig:long_walk} shows a very long (1.2km) walk using our controller, passing through different terrains without any human correction, except for manually controlling the robot's heading to chart its course.
Generally, position-based quadrupedal controllers perform well on various hard, unstructured surfaces. However, significant challenges remain when navigating soft or slippery terrain. As noted in~\cite{choi2023learning}, additional modifications to the controller or extensive domain randomization are often required to ensure robust real-world deployment on such surfaces. On the other hand, torque-based controllers leverage a more direct understanding of the robot's dynamics and terrain interactions, enabling a more robust response to those environments. The following section covers several challenging and out-of-distribution terrains, including height constraint, or soft and slippery grounds, showcasing the generalizability and adaptability of our approach.

\subsubsection{Squeezing into a Tunnel}
\begin{figure}[tbp]
    \centering
    \includegraphics[width=1\linewidth]{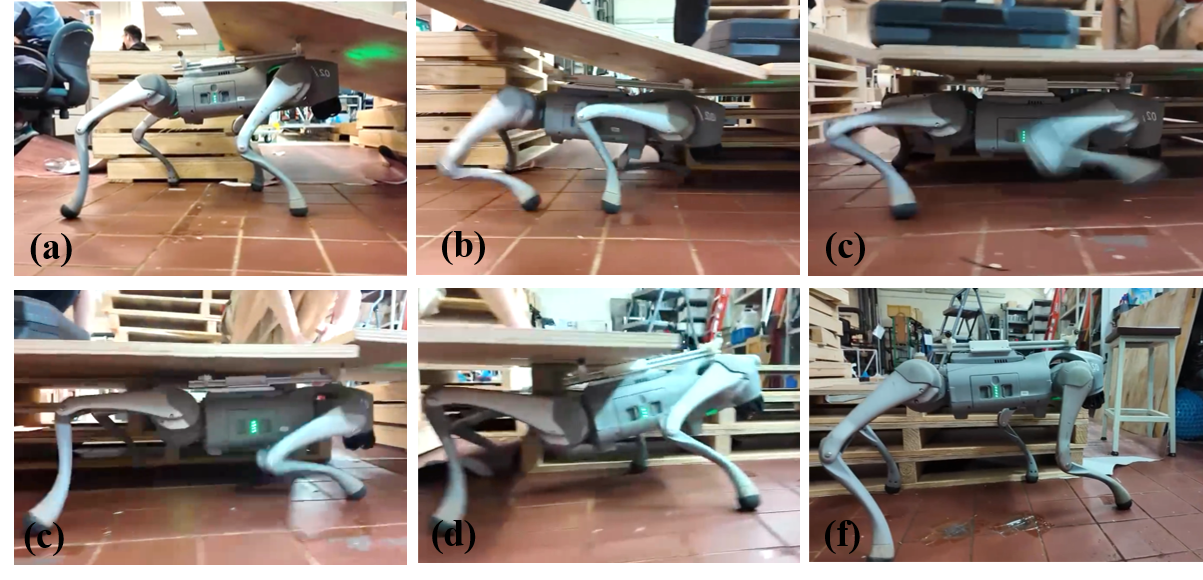}
    \vspace{-0.8cm}
    \caption{Locomotion through a height-constrained space. Notably, no height command or additional modifications were made to the robot or the policy, except for the addition of passive wheels on top of the robot's body to reduce friction from the tunnel ceiling.}
    \vspace{-0.5cm}
    \label{fig:tunnel_experiment}
\end{figure}

To evaluate the performance of our controller in height-constrained scenarios, we designed a tunnel traversal experiment in which the minimum tunnel height was approximately 30 cm. Unlike baseline methods which can hardly be compressed by vertical force, our approach is able to passively adapt to this situation. As shown in Fig.~\ref{fig:tunnel_experiment}, the robot was compressed to approximately half of its normal height after bumping into the tunnel ceiling, and was able to move forward while staying in this compressed state. Despite the fact that this scenario was never seen during training, and that no external height command was given, the controller autonomously adopted a crawling gait, allowing the robot to continue moving forward without getting stuck. Upon exiting the tunnel, the robot naturally returned to its upright posture and resumed its normal walking gait without requiring any external intervention.

\begin{figure}[t]
    \centering
    \includegraphics[width=1\linewidth]{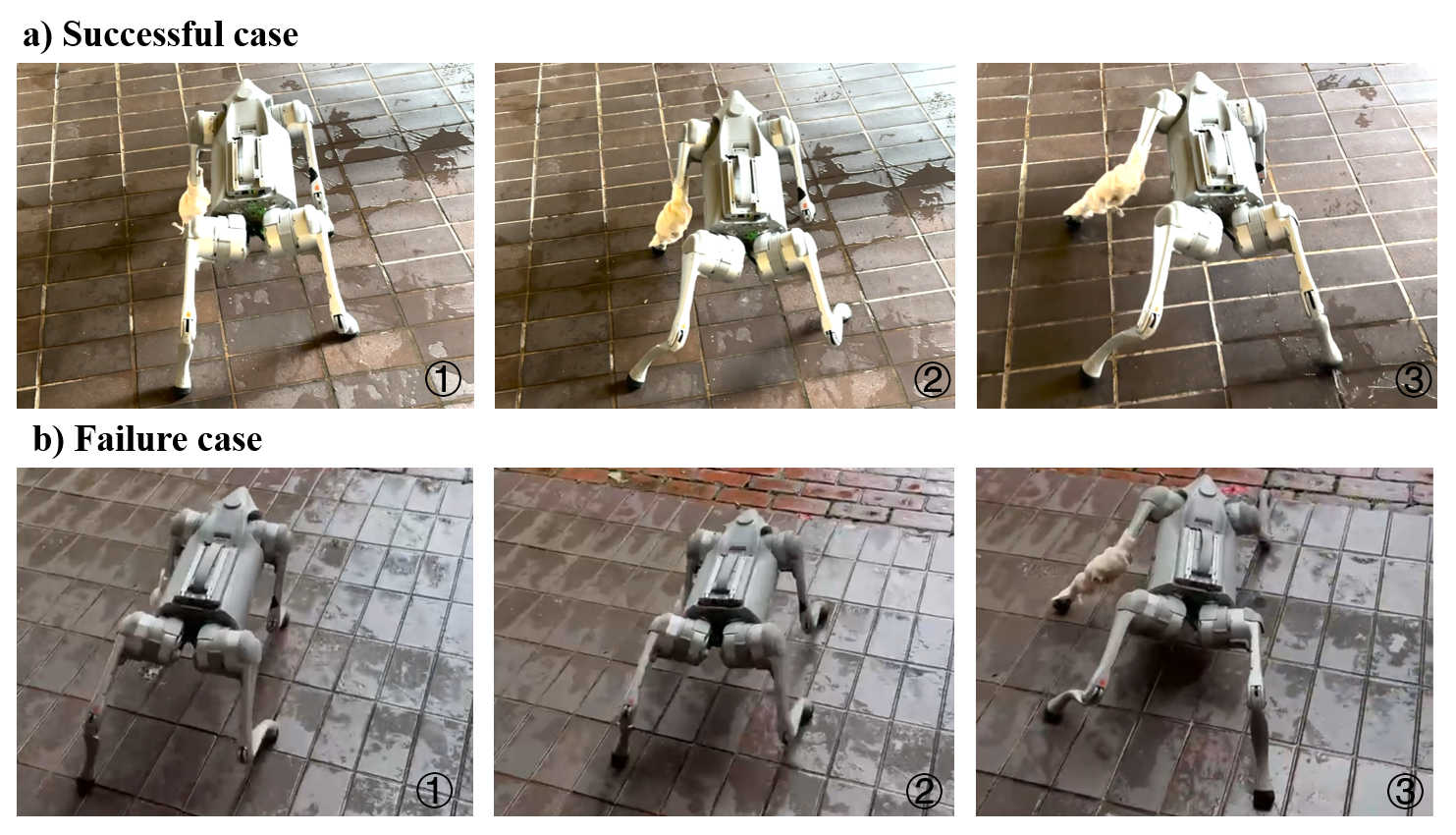}
    \vspace{-0.8cm}
    \caption{Locomotion on wet slippery surfaces, showing both success (a) and failure (b). Even when the foot of the robot slips and fall down in failure cases, the controller still does not exhibit wild, unsafe motions.}
    \label{fig:slip}
    \vspace{-0.7cm}
\end{figure}

\subsubsection{Slippery Surfaces}
Figure~\ref{fig:slip} demonstrates the robot's performance on a wet and slippery surface. In Figure~\ref{fig:slip}a, after experiencing a slip during locomotion, the robot quickly recovers its standing posture. Our controller dynamically adjusts the motor output to stabilize the robot without predefined recovery motions. In contrast, Figure~\ref{fig:slip}b shows a failure case, where the robot is given an abrupt command on the slippery surface. Note that the controller is still active with the robot applying torque through the thigh joints in an attempt to recover balance, yet no abrupt or flailing motion is produced. This behavior contrasts sharply with position-based controllers, where the thigh and calf joint would typically be retracted abruptly, often leading to tipping or other destabilizing outcomes. The torque-based controller's ability to exert gradual and compliant force enhances minimizes the aggressive reactions that could compromise stability.

\subsubsection{Soft Terrain}
\begin{figure}[tbp]
    \centering
    \includegraphics[width=1\linewidth]{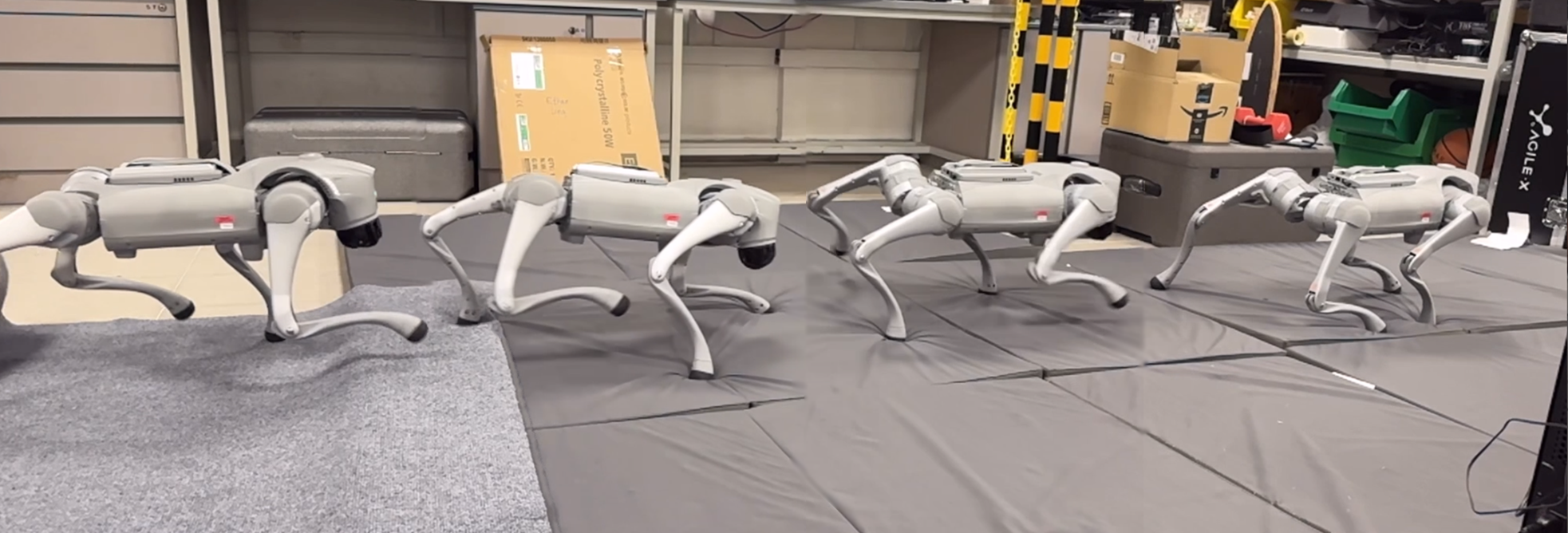}
    \vspace{-0.7cm}
    \caption{Locomotion on 10cm thick soft mattress with a velocity command of 0.8m/s. Our robot stops with the right most posture when the velocity command is finally set to 0.}
    \vspace{-0.5cm}
    \label{fig:deformable}
\end{figure}

In addition to slippery surfaces, deformable terrain presents significant challenges for position-based controllers. To evaluate performance, we conducted an experiment where the robot traversed a soft mattress approximately 10 cm thick. As shown in Figure~\ref{fig:deformable}, a position-based reinforcement learning controller struggled to cross the terrain effectively, often failing to maintain forward momentum or balance. In contrast, our torque-based controller demonstrated robust performance, successfully navigating the deformable surface without any additional modification. This highlights the advantage of torque control in dynamically adapting to terrain variability, leveraging compliant interaction with the environment for enhanced stability and maneuverability.

By directly regulating motor torques, our policy allows the robot to exhibit greater compliance and stability, reducing the dependency on domain randomization or manual tuning commonly needed for position-based approaches. We believe these advantages highlight the potential of torque-based control for robust real-world deployments in diverse environments.

\section{Limitations}
\label{sec:limitation}
\subsection{Challenges in Posture Maintenance}

Our current control strategy is torque-based, which offers enhanced compliance, enabling the robot to better adapt to impacts from various directions. However, this compliance also comes at the cost of posture maintenance. For instance, when carrying a medium-weight payload (5 kg), a scenario that position-based control systems can easily handle, our robot is not able to maintain its body height during walking.
This often results in the robot's calfs coming into contact with the ground, which risks damaging the robot. To address this limitation, we believe that further optimization of the biomechanical model may be necessary, to allow the robot to better understand and adapt (comply with, or safely resists to) different types of external forces. For example, introducing tunable parameters or introducing antagonistic muscle pairs could enable the robot to dynamically adjust the stiffness of its virtual muscles, thereby enhancing its posture maintenance capabilities when required.

\subsection{Gait Learning Limitations}

Compared to position-based control systems, which can learn and execute a diverse range of gaits and generate relatively more dynamic gaits, such as trot, amble, or gallop, our torque-based policy has so far only successfully learned a basic walking gait.
Even at high speeds, such as 2.0~m/s, the robot continues to utilize a walking gait pattern, merely increasing its stride length.
We believe that this limitation may hinder the robot's performance in highly dynamic scenarios and could lead to increased energy consumption.
In future work, we will seek ways to first manually guide the learning of different gaits, and then look for mechanisms that may allow them to emerge from the training (alongside methods to stably switch between them), once again by drawing inspiration from biological principles and structures.

\section{Conclusion}
\label{sec:conclusion}
In this work, we presented SATA, a bio-inspired torque-based learning framework for quadrupedal locomotion, aimed at achieving safer and more compliant robot behaviors. By incorporating biomechanical principles and adaptive learning mechanisms inspired by animal locomotion, SATA enhances compliance, adaptability, and generalization, enabling robots to interact more effectively with diverse and challenging environments.
Our framework leverages a muscle-like model to smoothen motion commands and mitigate suboptimal behaviors, alongside a growth-inspired training mechanism that progressively unlocks robot capabilities and improves early-stage exploration efficiency. Experimental results demonstrate SATA's robustness in handling a variety of scenarios, including unseen velocity commands, soft or slippery terrains, and real-world disturbances such as sudden torque limitations and external impacts.
Moreover, SATA achieves zero-shot sim-to-real transfer, removing the need for additional hardware fine-tuning while maintaining safe and robust locomotion. 

Our future work will extend SATA's capabilities to more complex terrains and agile behaviors, further improving its adaptability in unstructured/dynamic environments.
We further plan to explore hybrid learning strategies that integrate adaptive muscle-like actuation, to further boost compliance and robustness.
We envision that these advancements will bring us ever closer to deploying safe and versatile torque-based controllers in real-world scenarios.

\section{ACKNOWLEDGMENTS}
This work was supported by the Singapore Ministry of Education Academic Research Fund Tier 1, as well as the National Research Foundation, Singapore (NRF), the Maritime and Port Authority of Singapore (MPA), the Singapore Maritime Institute (SMI) under its Maritime Transformation Programme (Project No. SMI-2022-MTP-01), and the Swiss National Science Foundation, under Grant 200021\_197237. 

\bibliographystyle{unsrtnat}
\bibliography{references}

\begin{thebibliography}{75}
\providecommand{\natexlab}[1]{#1}
\providecommand{\url}[1]{\texttt{#1}}
\expandafter\ifx\csname urlstyle\endcsname\relax
  \providecommand{\doi}[1]{doi: #1}\else
  \providecommand{\doi}{doi: \begingroup \urlstyle{rm}\Url}\fi

\bibitem[Luo et~al.(2020)Luo, Soeseno, Chen, and Chen]{luo2020carl}
Ying-Sheng Luo, Jonathan~Hans Soeseno, Trista Pei-Chun Chen, and Wei-Chao Chen.
\newblock Carl: Controllable agent with reinforcement learning for quadruped
  locomotion.
\newblock \emph{ACM Transactions on Graphics (TOG)}, 39\penalty0 (4):\penalty0
  38--1, 2020.

\bibitem[Tan et~al.(2018)Tan, Zhang, Coumans, Iscen, Bai, Hafner, Bohez, and
  Vanhoucke]{tan2018sim}
Jie Tan, Tingnan Zhang, Erwin Coumans, Atil Iscen, Yunfei Bai, Danijar Hafner,
  Steven Bohez, and Vincent Vanhoucke.
\newblock Sim-to-real: Learning agile locomotion for quadruped robots.
\newblock \emph{arXiv preprint arXiv:1804.10332}, 2018.

\bibitem[Choi et~al.(2023)Choi, Ji, Park, Kim, Mun, Lee, and
  Hwangbo]{choi2023learning}
Suyoung Choi, Gwanghyeon Ji, Jeongsoo Park, Hyeongjun Kim, Juhyeok Mun,
  Jeong~Hyun Lee, and Jemin Hwangbo.
\newblock Learning quadrupedal locomotion on deformable terrain.
\newblock \emph{Science Robotics}, 8\penalty0 (74):\penalty0 eade2256, 2023.

\bibitem[Lee et~al.(2020)Lee, Hwangbo, Wellhausen, Koltun, and
  Hutter]{lee2020learning}
Joonho Lee, Jemin Hwangbo, Lorenz Wellhausen, Vladlen Koltun, and Marco Hutter.
\newblock Learning quadrupedal locomotion over challenging terrain.
\newblock \emph{Science robotics}, 5\penalty0 (47):\penalty0 eabc5986, 2020.

\bibitem[Miki et~al.(2022)Miki, Lee, Hwangbo, Wellhausen, Koltun, and
  Hutter]{miki2022learning}
Takahiro Miki, Joonho Lee, Jemin Hwangbo, Lorenz Wellhausen, Vladlen Koltun,
  and Marco Hutter.
\newblock Learning robust perceptive locomotion for quadrupedal robots in the
  wild.
\newblock \emph{Science robotics}, 7\penalty0 (62):\penalty0 eabk2822, 2022.

\bibitem[Hu et~al.(2019)Hu, Shao, Cao, Xiao, Li, and Ma]{hu2019learning}
Biao Hu, Shibo Shao, Zhengcai Cao, Qing Xiao, Qunzhi Li, and Chao Ma.
\newblock Learning a faster locomotion gait for a quadruped robot with
  model-free deep reinforcement learning.
\newblock In \emph{2019 IEEE International Conference on Robotics and
  Biomimetics (ROBIO)}, pages 1097--1102. IEEE, 2019.

\bibitem[Tsounis et~al.(2020)Tsounis, Alge, Lee, Farshidian, and
  Hutter]{tsounis2020deepgait}
Vassilios Tsounis, Mitja Alge, Joonho Lee, Farbod Farshidian, and Marco Hutter.
\newblock Deepgait: Planning and control of quadrupedal gaits using deep
  reinforcement learning.
\newblock \emph{IEEE Robotics and Automation Letters}, 5\penalty0 (2):\penalty0
  3699--3706, 2020.

\bibitem[Kim and Lee(2021)]{kim2021quadruped}
Taehei Kim and Sung-Hee Lee.
\newblock Quadruped locomotion on non-rigid terrain using reinforcement
  learning.
\newblock \emph{arXiv preprint arXiv:2107.02955}, 2021.

\bibitem[Margolis and Agrawal(2023)]{margolis2023walk}
Gabriel~B Margolis and Pulkit Agrawal.
\newblock Walk these ways: Tuning robot control for generalization with
  multiplicity of behavior.
\newblock In \emph{Conference on Robot Learning}, pages 22--31. PMLR, 2023.

\bibitem[Sombolestan and Nguyen(2024)]{sombolestan2024adaptive}
Mohsen Sombolestan and Quan Nguyen.
\newblock Adaptive force-based control of dynamic legged locomotion over uneven
  terrain.
\newblock \emph{IEEE Transactions on Robotics}, 2024.

\bibitem[Sood et~al.(2024)Sood, Sun, Li, and Sartoretti]{sood2024decap}
Shivam Sood, Ge~Sun, Peizhuo Li, and Guillaume Sartoretti.
\newblock Decap: Decaying action priors for accelerated imitation learning of
  torque-based legged locomotion policies.
\newblock In \emph{2024 IEEE/RSJ International Conference on Intelligent Robots
  and Systems (IROS)}, pages 2809--2815. IEEE, 2024.

\bibitem[Chen et~al.(2023)Chen, Zhang, Mueller, Rai, and
  Sreenath]{chen2023learning}
Shuxiao Chen, Bike Zhang, Mark~W Mueller, Akshara Rai, and Koushil Sreenath.
\newblock Learning torque control for quadrupedal locomotion.
\newblock In \emph{2023 IEEE-RAS 22nd International Conference on Humanoid
  Robots (Humanoids)}, pages 1--8. IEEE, 2023.

\bibitem[Seow(2013)]{seow2013hill}
Chun~Y Seow.
\newblock Hill’s equation of muscle performance and its hidden insight on
  molecular mechanisms.
\newblock \emph{Journal of General Physiology}, 142\penalty0 (6):\penalty0
  561--573, 2013.

\bibitem[Enoka and Duchateau(2008)]{enoka2008muscle}
Roger~M Enoka and Jacques Duchateau.
\newblock Muscle fatigue: what, why and how it influences muscle function.
\newblock \emph{The Journal of physiology}, 586\penalty0 (1):\penalty0 11--23,
  2008.

\bibitem[Bellegarda et~al.(2022)Bellegarda, Chen, Liu, and
  Nguyen]{bellegarda2022robust}
Guillaume Bellegarda, Yiyu Chen, Zhuochen Liu, and Quan Nguyen.
\newblock Robust high-speed running for quadruped robots via deep reinforcement
  learning.
\newblock In \emph{2022 IEEE/RSJ International Conference on Intelligent Robots
  and Systems (IROS)}, pages 10364--10370. IEEE, 2022.

\bibitem[Bellegarda et~al.(2024)Bellegarda, Nguyen, and
  Nguyen]{bellegarda2024robust}
Guillaume Bellegarda, Chuong Nguyen, and Quan Nguyen.
\newblock Robust quadruped jumping via deep reinforcement learning.
\newblock \emph{Robotics and Autonomous Systems}, 182:\penalty0 104799, 2024.

\bibitem[Ren et~al.(2021)Ren, Wang, Yang, and Cao]{ren2021learning}
Liang Ren, Chunlei Wang, Ya~Yang, and Zhiqiang Cao.
\newblock A learning-based control approach for blind quadrupedal locomotion
  with guided-drl and hierarchical-drl.
\newblock In \emph{2021 IEEE International Conference on Robotics and
  Biomimetics (ROBIO)}, pages 881--886. IEEE, 2021.

\bibitem[Ding et~al.(2019)Ding, Pandala, and Park]{ding2019real}
Yanran Ding, Abhishek Pandala, and Hae-Won Park.
\newblock Real-time model predictive control for versatile dynamic motions in
  quadrupedal robots.
\newblock In \emph{2019 International Conference on Robotics and Automation
  (ICRA)}, pages 8484--8490. IEEE, 2019.

\bibitem[Kim et~al.(2019)Kim, Di~Carlo, Katz, Bledt, and Kim]{kim2019highly}
Donghyun Kim, Jared Di~Carlo, Benjamin Katz, Gerardo Bledt, and Sangbae Kim.
\newblock Highly dynamic quadruped locomotion via whole-body impulse control
  and model predictive control.
\newblock \emph{arXiv preprint arXiv:1909.06586}, 2019.

\bibitem[Neunert et~al.(2018)Neunert, St{\"a}uble, Giftthaler, Bellicoso,
  Carius, Gehring, Hutter, and Buchli]{neunert2018whole}
Michael Neunert, Markus St{\"a}uble, Markus Giftthaler, Carmine~D Bellicoso,
  Jan Carius, Christian Gehring, Marco Hutter, and Jonas Buchli.
\newblock Whole-body nonlinear model predictive control through contacts for
  quadrupeds.
\newblock \emph{IEEE Robotics and Automation Letters}, 3\penalty0 (3):\penalty0
  1458--1465, 2018.

\bibitem[Grandia et~al.(2023)Grandia, Jenelten, Yang, Farshidian, and
  Hutter]{grandia2023perceptive}
Ruben Grandia, Fabian Jenelten, Shaohui Yang, Farbod Farshidian, and Marco
  Hutter.
\newblock Perceptive locomotion through nonlinear model-predictive control.
\newblock \emph{IEEE Transactions on Robotics}, 39\penalty0 (5):\penalty0
  3402--3421, 2023.

\bibitem[Peng et~al.(2018)Peng, Abbeel, Levine, and Van~de
  Panne]{peng2018deepmimic}
Xue~Bin Peng, Pieter Abbeel, Sergey Levine, and Michiel Van~de Panne.
\newblock Deepmimic: Example-guided deep reinforcement learning of
  physics-based character skills.
\newblock \emph{ACM Transactions On Graphics (TOG)}, 37\penalty0 (4):\penalty0
  1--14, 2018.

\bibitem[Hwangbo et~al.(2019)Hwangbo, Lee, Dosovitskiy, Bellicoso, Tsounis,
  Koltun, and Hutter]{hwangbo2019learning}
Jemin Hwangbo, Joonho Lee, Alexey Dosovitskiy, Dario Bellicoso, Vassilios
  Tsounis, Vladlen Koltun, and Marco Hutter.
\newblock Learning agile and dynamic motor skills for legged robots.
\newblock \emph{Science Robotics}, 4\penalty0 (26):\penalty0 eaau5872, 2019.

\bibitem[Aractingi et~al.(2023)Aractingi, L{\'e}ziart, Flayols, Perez,
  Silander, and Sou{\`e}res]{aractingi2023controlling}
Michel Aractingi, Pierre-Alexandre L{\'e}ziart, Thomas Flayols, Julien Perez,
  Tomi Silander, and Philippe Sou{\`e}res.
\newblock Controlling the solo12 quadruped robot with deep reinforcement
  learning.
\newblock \emph{scientific Reports}, 13\penalty0 (1):\penalty0 11945, 2023.

\bibitem[Yang et~al.(2020)Yang, Caluwaerts, Iscen, Zhang, Tan, and
  Sindhwani]{yang2020data}
Yuxiang Yang, Ken Caluwaerts, Atil Iscen, Tingnan Zhang, Jie Tan, and Vikas
  Sindhwani.
\newblock Data efficient reinforcement learning for legged robots.
\newblock In \emph{Conference on Robot Learning}, pages 1--10. PMLR, 2020.

\bibitem[Allshire et~al.(2021)Allshire, Mart{\'\i}n-Mart{\'\i}n, Lin, Manuel,
  Savarese, and Garg]{allshire2021laser}
Arthur Allshire, Roberto Mart{\'\i}n-Mart{\'\i}n, Charles Lin, Shawn Manuel,
  Silvio Savarese, and Animesh Garg.
\newblock Laser: Learning a latent action space for efficient reinforcement
  learning.
\newblock In \emph{2021 IEEE International Conference on Robotics and
  Automation (ICRA)}, pages 6650--6656. IEEE, 2021.

\bibitem[Kim et~al.(2023{\natexlab{a}})Kim, Kim, and Park]{kim2023automated}
MyeongSeop Kim, Jung-Su Kim, and Jae-Han Park.
\newblock Automated hyperparameter tuning in reinforcement learning for
  quadrupedal robot locomotion.
\newblock \emph{Electronics}, 13\penalty0 (1):\penalty0 116,
  2023{\natexlab{a}}.

\bibitem[Lyu et~al.(2024)Lyu, Lang, Zhao, Zhang, Ding, and Wang]{lyu2024rl2ac}
Shangke Lyu, Xin Lang, Han Zhao, Hongyin Zhang, Pengxiang Ding, and Donglin
  Wang.
\newblock Rl2ac: Reinforcement learning-based rapid online adaptive control for
  legged robot robust locomotion.
\newblock In \emph{Proceedings of the Robotics: Science and Systems}, 2024.

\bibitem[Buchli et~al.(2009)Buchli, Kalakrishnan, Mistry, Pastor, and
  Schaal]{buchli2009compliant}
Jonas Buchli, Mrinal Kalakrishnan, Michael Mistry, Peter Pastor, and Stefan
  Schaal.
\newblock Compliant quadruped locomotion over rough terrain.
\newblock In \emph{2009 IEEE/RSJ international conference on Intelligent robots
  and systems}, pages 814--820. IEEE, 2009.

\bibitem[Calanca et~al.(2015)Calanca, Muradore, and Fiorini]{calanca2015review}
Andrea Calanca, Riccardo Muradore, and Paolo Fiorini.
\newblock A review of algorithms for compliant control of stiff and
  fixed-compliance robots.
\newblock \emph{IEEE/ASME transactions on mechatronics}, 21\penalty0
  (2):\penalty0 613--624, 2015.

\bibitem[Kim et~al.(2023{\natexlab{b}})Kim, Berseth, Schwartz, and
  Park]{kim2023torque}
Donghyeon Kim, Glen Berseth, Mathew Schwartz, and Jaeheung Park.
\newblock Torque-based deep reinforcement learning for task-and-robot agnostic
  learning on bipedal robots using sim-to-real transfer.
\newblock \emph{IEEE Robotics and Automation Letters}, 2023{\natexlab{b}}.

\bibitem[Makoviychuk et~al.(2021)Makoviychuk, Wawrzyniak, Guo, Lu, Storey,
  Macklin, Hoeller, Rudin, Allshire, Handa, et~al.]{makoviychuk2021isaac}
Viktor Makoviychuk, Lukasz Wawrzyniak, Yunrong Guo, Michelle Lu, Kier Storey,
  Miles Macklin, David Hoeller, Nikita Rudin, Arthur Allshire, Ankur Handa,
  et~al.
\newblock Isaac gym: High performance gpu-based physics simulation for robot
  learning.
\newblock \emph{arXiv preprint arXiv:2108.10470}, 2021.

\bibitem[Rudin et~al.(2022{\natexlab{a}})Rudin, Hoeller, Reist, and
  Hutter]{rudin2022learning}
Nikita Rudin, David Hoeller, Philipp Reist, and Marco Hutter.
\newblock Learning to walk in minutes using massively parallel deep
  reinforcement learning.
\newblock In \emph{Conference on Robot Learning}, pages 91--100. PMLR,
  2022{\natexlab{a}}.

\bibitem[Rudin et~al.(2022{\natexlab{b}})Rudin, Hoeller, Reist, and
  Hutter]{rudin2022learningwalkminutesusing}
Nikita Rudin, David Hoeller, Philipp Reist, and Marco Hutter.
\newblock Learning to walk in minutes using massively parallel deep
  reinforcement learning, 2022{\natexlab{b}}.
\newblock URL \url{https://arxiv.org/abs/2109.11978}.

\bibitem[Zuo et~al.(2024)Zuo, Wang, Gong, and Yu]{zuo2024learning}
Guoyu Zuo, Yong Wang, Daoxiong Gong, and Shuangyue Yu.
\newblock Learning quadrupedal locomotion on tough terrain using an asymmetric
  terrain feature mining network.
\newblock \emph{Applied Intelligence}, 54\penalty0 (22):\penalty0 11547--11563,
  2024.

\bibitem[Chen and Nguyen(2024)]{chen2024learning}
Yiyu Chen and Quan Nguyen.
\newblock Learning agile locomotion and adaptive behaviors via rl-augmented
  mpc.
\newblock In \emph{2024 IEEE International Conference on Robotics and
  Automation (ICRA)}, pages 11436--11442. IEEE, 2024.

\bibitem[Kumar et~al.(2021)Kumar, Fu, Pathak, and Malik]{kumar2021rma}
Ashish Kumar, Zipeng Fu, Deepak Pathak, and Jitendra Malik.
\newblock Rma: Rapid motor adaptation for legged robots.
\newblock \emph{arXiv preprint arXiv:2107.04034}, 2021.

\bibitem[Gangapurwala et~al.(2022)Gangapurwala, Geisert, Orsolino, Fallon, and
  Havoutis]{gangapurwala2022rloc}
Siddhant Gangapurwala, Mathieu Geisert, Romeo Orsolino, Maurice Fallon, and
  Ioannis Havoutis.
\newblock Rloc: Terrain-aware legged locomotion using reinforcement learning
  and optimal control.
\newblock \emph{IEEE Transactions on Robotics}, 38\penalty0 (5):\penalty0
  2908--2927, 2022.

\bibitem[Yao et~al.(2021)Yao, Wang, Wang, Yang, Zhang, Wang, and
  Wu]{yao2021hierarchical}
Qingfeng Yao, Jilong Wang, Donglin Wang, Shuyu Yang, Hongyin Zhang, Yinuo Wang,
  and Zhengqing Wu.
\newblock Hierarchical terrain-aware control for quadrupedal locomotion by
  combining deep reinforcement learning and optimal control.
\newblock In \emph{2021 IEEE/RSJ International Conference on Intelligent Robots
  and Systems (IROS)}, pages 4546--4551. IEEE, 2021.

\bibitem[Zhang et~al.(2022{\natexlab{a}})Zhang, An, Wei, and
  Ma]{zhang2022learning}
Zhitong Zhang, Honglei An, Qing Wei, and Hongxu Ma.
\newblock Learning-based model predictive control for quadruped locomotion on
  slippery ground.
\newblock In \emph{2022 4th International Conference on Control and Robotics
  (ICCR)}, pages 47--52. IEEE, 2022{\natexlab{a}}.

\bibitem[Zhang et~al.(2022{\natexlab{b}})Zhang, Chang, Ma, An, and
  Lang]{zhang2022model}
Zhitong Zhang, Xu~Chang, Hongxu Ma, Honglei An, and Lin Lang.
\newblock Model predictive control of quadruped robot based on reinforcement
  learning.
\newblock \emph{Applied Sciences}, 13\penalty0 (1):\penalty0 154,
  2022{\natexlab{b}}.

\bibitem[Abdi et~al.(2019)Abdi, Malakoutian, Oxland, and
  Fels]{abdi2019reinforcement}
Amir~H Abdi, Masoud Malakoutian, Thomas Oxland, and Sidney Fels.
\newblock Reinforcement learning for high-dimensional continuous control in
  biomechanics: an intro to artisynth-rl.
\newblock \emph{arXiv preprint arXiv:1910.13859}, 2019.

\bibitem[Ouyang et~al.(2021)Ouyang, Chi, Pang, Liang, and
  Ren]{ouyang2021adaptive}
Wenjuan Ouyang, Haozhen Chi, Jiangnan Pang, Wenyu Liang, and Qinyuan Ren.
\newblock Adaptive locomotion control of a hexapod robot via bio-inspired
  learning.
\newblock \emph{Frontiers in Neurorobotics}, 15:\penalty0 627157, 2021.

\bibitem[Humphreys and Zhou(2024)]{humphreys2024learning}
Joseph Humphreys and Chengxu Zhou.
\newblock Learning to adapt: Bio-inspired gait strategies for versatile
  quadruped locomotion.
\newblock \emph{arXiv preprint arXiv:2412.09440}, 2024.

\bibitem[Wang et~al.(2021)Wang, Hu, and Zhu]{wang2021cpg}
Jiayu Wang, Chuxiong Hu, and Yu~Zhu.
\newblock Cpg-based hierarchical locomotion control for modular quadrupedal
  robots using deep reinforcement learning.
\newblock \emph{IEEE Robotics and Automation Letters}, 6\penalty0 (4):\penalty0
  7193--7200, 2021.

\bibitem[Wei and Webb(2018)]{wei2018bio}
Tianqi Wei and Barbara Webb.
\newblock A bio-inspired reinforcement learning rule to optimise dynamical
  neural networks for robot control.
\newblock In \emph{2018 IEEE/RSJ International Conference on Intelligent Robots
  and Systems (IROS)}, pages 556--561. IEEE, 2018.

\bibitem[Zhang et~al.(2023)Zhang, Xiao, Zhang, and Pan]{zhang2023synloco}
Xinyu Zhang, Zhiyuan Xiao, Qingrui Zhang, and Wei Pan.
\newblock Synloco: Synthesizing central pattern generator and reinforcement
  learning for quadruped locomotion.
\newblock \emph{arXiv preprint arXiv:2310.06606}, 2023.

\bibitem[Peng et~al.(2020)Peng, Coumans, Zhang, Lee, Tan, and
  Levine]{peng2020learning}
Xue~Bin Peng, Erwin Coumans, Tingnan Zhang, Tsang-Wei Lee, Jie Tan, and Sergey
  Levine.
\newblock Learning agile robotic locomotion skills by imitating animals.
\newblock \emph{arXiv preprint arXiv:2004.00784}, 2020.

\bibitem[Fu et~al.(2021)Fu, Kumar, Malik, and Pathak]{fu2021minimizing}
Zipeng Fu, Ashish Kumar, Jitendra Malik, and Deepak Pathak.
\newblock Minimizing energy consumption leads to the emergence of gaits in
  legged robots.
\newblock \emph{arXiv preprint arXiv:2111.01674}, 2021.

\bibitem[Bellegarda and Ijspeert(2022)]{bellegarda2022cpg}
Guillaume Bellegarda and Auke Ijspeert.
\newblock {CPG-RL}: Learning central pattern generators for quadruped
  locomotion.
\newblock \emph{IEEE Robotics and Automation Letters}, 7\penalty0 (4):\penalty0
  12547--12554, 2022.

\bibitem[Sun et~al.(2024)Sun, Shafiee, Li, Bellegarda, Ijspeert, and
  Sartoretti]{sun2024learning}
Ge~Sun, Milad Shafiee, Peizhuo Li, Guillaume Bellegarda, Auke Ijspeert, and
  Guillaume Sartoretti.
\newblock Learning-based hierarchical control: Emulating the central nervous
  system for bio-inspired legged robot locomotion.
\newblock \emph{arXiv preprint arXiv:2404.17815}, 2024.

\bibitem[He et~al.(2024)He, Lei, Ze, Sreenath, Li, and Xu]{he2024learning}
Zhengmao He, Kun Lei, Yanjie Ze, Koushil Sreenath, Zhongyu Li, and Huazhe Xu.
\newblock Learning visual quadrupedal loco-manipulation from demonstrations.
\newblock \emph{arXiv preprint arXiv:2403.20328}, 2024.

\bibitem[Jain et~al.(2019)Jain, Iscen, and Caluwaerts]{jain2019hierarchical}
Deepali Jain, Atil Iscen, and Ken Caluwaerts.
\newblock Hierarchical reinforcement learning for quadruped locomotion.
\newblock In \emph{2019 IEEE/RSJ International Conference on Intelligent Robots
  and Systems (IROS)}, pages 7551--7557. IEEE, 2019.

\bibitem[Jain et~al.(2020)Jain, Iscen, and Caluwaerts]{jain2020pixels}
Deepali Jain, Atil Iscen, and Ken Caluwaerts.
\newblock From pixels to legs: Hierarchical learning of quadruped locomotion.
\newblock \emph{arXiv preprint arXiv:2011.11722}, 2020.

\bibitem[Botterman and Gonyea(1986)]{botterman1986gradation}
abc Botterman and abc Gonyea.
\newblock Gradation of isometric tension by different activation rates in motor
  units of cat flexor carpi radialis muscle.
\newblock \emph{Journal of neurophysiology}, 56\penalty0 (2):\penalty0
  494--506, 1986.

\bibitem[Caillet et~al.(2023)Caillet, Phillips, Farina, and
  Modenese]{caillet2023motoneuron}
Arnault~H Caillet, Andrew~TM Phillips, Dario Farina, and Luca Modenese.
\newblock Motoneuron-driven computational muscle modelling with motor unit
  resolution and subject-specific musculoskeletal anatomy.
\newblock \emph{PLOS Computational Biology}, 19\penalty0 (12):\penalty0
  e1011606, 2023.

\bibitem[Callahan et~al.(2013)Callahan, Umberger, and
  Kent-Braun]{callahan2013computational}
Damien~M Callahan, Brian~R Umberger, and Jane~A Kent-Braun.
\newblock A computational model of torque generation: neural, contractile,
  metabolic and musculoskeletal components.
\newblock \emph{PloS one}, 8\penalty0 (2):\penalty0 e56013, 2013.

\bibitem[Hassani et~al.(2014)Hassani, Tjahjowidodo, and Do]{hassani2014survey}
Vahid Hassani, Tegoeh Tjahjowidodo, and Thanh~Nho Do.
\newblock A survey on hysteresis modeling, identification and control.
\newblock \emph{Mechanical systems and signal processing}, 49\penalty0
  (1-2):\penalty0 209--233, 2014.

\bibitem[Schmitt et~al.(2019)Schmitt, G{\"u}nther, and
  H{\"a}ufle]{schmitt2019dynamics}
Syn Schmitt, Michael G{\"u}nther, and Daniel~FB H{\"a}ufle.
\newblock The dynamics of the skeletal muscle: A systems biophysics perspective
  on muscle modeling with the focus on hill-type muscle models.
\newblock \emph{GAMM-Mitteilungen}, 42\penalty0 (3):\penalty0 e201900013, 2019.

\bibitem[Romero and Alonso(2016)]{romero2016comparison}
F~Romero and FJ~Alonso.
\newblock A comparison among different hill-type contraction dynamics
  formulations for muscle force estimation.
\newblock \emph{Mechanical Sciences}, 7\penalty0 (1):\penalty0 19--29, 2016.

\bibitem[Till et~al.(2008)Till, Siebert, Rode, and
  Blickhan]{till2008characterization}
Olaf Till, Tobias Siebert, Christian Rode, and Reinhard Blickhan.
\newblock Characterization of isovelocity extension of activated muscle: a
  hill-type model for eccentric contractions and a method for parameter
  determination.
\newblock \emph{Journal of theoretical biology}, 255\penalty0 (2):\penalty0
  176--187, 2008.

\bibitem[Boonstra and Beek(2008)]{boonstra2008fatigue}
TW~Boonstra and PJ~Beek.
\newblock Fatigue-related changes in motor-unit synchronization of quadriceps
  muscles within and across legs.
\newblock \emph{Journal of Electromyography and Kinesiology}, 18\penalty0
  (5):\penalty0 717--731, 2008.

\bibitem[Cowley and Gates(2017)]{cowley2017inter}
Jeffrey~C Cowley and Deanna~H Gates.
\newblock Inter-joint coordination changes during and after muscle fatigue.
\newblock \emph{Human Movement Science}, 56:\penalty0 109--118, 2017.

\bibitem[Qin et~al.(2019)Qin, Gao, and Bai]{qin2019sim}
Bangyu Qin, Yue Gao, and Yi~Bai.
\newblock Sim-to-real: Six-legged robot control with deep reinforcement
  learning and curriculum learning.
\newblock In \emph{2019 4th International Conference on Robotics and Automation
  Engineering (ICRAE)}, pages 1--5. IEEE, 2019.

\bibitem[Yu et~al.(2018)Yu, Turk, and Liu]{yu2018learning}
Wenhao Yu, Greg Turk, and C~Karen Liu.
\newblock Learning symmetric and low-energy locomotion.
\newblock \emph{ACM Transactions on Graphics (TOG)}, 37\penalty0 (4):\penalty0
  1--12, 2018.

\bibitem[Li et~al.(2024)Li, Wang, Pang, Bai, Hu, Liu, Wang, and
  Li]{li2024learning}
Sicen Li, Gang Wang, Yiming Pang, Panju Bai, Shihao Hu, Zhaojin Liu, Liquan
  Wang, and Jiawei Li.
\newblock Learning agility and adaptive legged locomotion via curricular
  hindsight reinforcement learning.
\newblock \emph{Scientific Reports}, 14\penalty0 (1):\penalty0 28089, 2024.

\bibitem[Kobayashi and Sugino(2020)]{kobayashi2020reinforcement}
Taisuke Kobayashi and Toshiki Sugino.
\newblock Reinforcement learning for quadrupedal locomotion with design of
  continual--hierarchical curriculum.
\newblock \emph{Engineering Applications of Artificial Intelligence},
  95:\penalty0 103869, 2020.

\bibitem[Berseth et~al.(2018)Berseth, Xie, Cernek, and Van~de
  Panne]{berseth2018progressive}
Glen Berseth, Cheng Xie, Paul Cernek, and Michiel Van~de Panne.
\newblock Progressive reinforcement learning with distillation for
  multi-skilled motion control.
\newblock \emph{arXiv preprint arXiv:1802.04765}, 2018.

\bibitem[Li et~al.(2023)Li, Tian, Tong, Niu, and Liu]{li2023robust}
Yike Li, Yunzhe Tian, Endong Tong, Wenjia Niu, and Jiqiang Liu.
\newblock Robust reinforcement learning via progressive task sequence.
\newblock In \emph{IJCAI}, pages 455--463, 2023.

\bibitem[Riedmiller and Springenberg(2018)]{riedmiller2018learning}
Martin Riedmiller and Jost Springenberg.
\newblock Learning by playing solving sparse reward tasks from scratch.
\newblock In \emph{International conference on machine learning}, pages
  4344--4353. PMLR, 2018.

\bibitem[Faust and Mehta(2019)]{faust2019evolving}
Aleksandra Faust and Dar Mehta.
\newblock Evolving rewards to automate reinforcement learning.
\newblock \emph{arXiv preprint arXiv:1905.07628}, 2019.

\bibitem[Ardelean and Suteu(2005)]{ardelean2005estimation}
A~Ardelean and E~Suteu.
\newblock The estimation of the growth curve at dog.
\newblock \emph{Bull Univ Agric Sci Vet Med Cluj-Napoca-Vet Med}, 63:\penalty0
  175--181, 2005.

\bibitem[Larsson and Karlsson(1979)]{larsson1979muscle}
Lars Larsson and Jan Karlsson.
\newblock Muscle strength and speed of movement in relation to age and muscle
  morphology.
\newblock \emph{Journal of Applied Physiology}, 46\penalty0 (3):\penalty0
  451--456, 1979.

\bibitem[FH~Gage(2004)]{gage2004structural}
abc FH~Gage.
\newblock Structural plasticity of the adult brain.
\newblock \emph{Dialogues in clinical neuroscience}, 6\penalty0 (2):\penalty0
  135--141, 2004.

\bibitem[Stampanoni~Bassi and Buttari(2019)]{stampanoni2019synaptic}
abc Stampanoni~Bassi and abc Buttari.
\newblock Synaptic plasticity shapes brain connectivity: implications for
  network topology.
\newblock \emph{International journal of molecular sciences}, 20\penalty0
  (24):\penalty0 6193, 2019.

\end{thebibliography}
\end{document}